\documentclass{article}

\usepackage[preprint]{neurips_2024}
\setcitestyle{numbers,square,citesep={,}}

\usepackage[utf8]{inputenc} 
\usepackage[T1]{fontenc}    
\usepackage{hyperref}       
\usepackage{url}            
\usepackage{booktabs}       
\usepackage{amsfonts}       
\usepackage{nicefrac}       
\usepackage{microtype}      
\usepackage{xcolor}         
\usepackage{multirow}

\usepackage{wrapfig}
\usepackage{adjustbox}
\usepackage{graphicx}
\usepackage{tikz}
\usepackage{comment}
\usepackage{amsmath,amssymb} 
\usepackage{color}
\usepackage{float}
\usepackage{subcaption}
\usepackage{pdflscape}
\usepackage{lipsum}

\usepackage{pifont}
\newcommand{\cmark}{\ding{51}}%
\newcommand{\xmark}{\ding{55}}%

\newcommand{\huien}[1]{\textcolor{black}{#1}}
\newcommand{\newhuien}[1]{\textcolor{black}{#1}}
\newcommand{\lei}[1]{\textcolor{black}{#1}}

\usepackage{hyperref}
\hypersetup{
  colorlinks   = true,    
  urlcolor     = blue,    
  linkcolor    = blue,    
  citecolor    = blue      
}

\title{Disco4D: Disentangled 4D Human Generation and Animation from a Single Image}

\author{
    Hui En Pang\textsuperscript{1}, \quad
    Shuai Liu\textsuperscript{3}, \quad
    Zhongang Cai\textsuperscript{1,2,3}, \quad
    Lei Yang\textsuperscript{2,3}, \quad
    \\
    \textbf{Tianwei Zhang\textsuperscript{1}}, \quad
    \textbf{Ziwei Liu\textsuperscript{1}} \\
    \textsuperscript{1}S-Lab, Nanyang Technological University \quad
    \textsuperscript{2}SenseTime Research \quad
    \textsuperscript{3}Shanghai AI Laboratory \quad \\
    \texttt{\{huien001, tianwei.zhang, ziwei.liu\}@ntu.edu.sg} \quad \\ \texttt{\{caizhongang, yanglei\}@sensetime.com} \quad \\ \texttt{\{liushuai\}@pjlab.org.cn}
}

\begin{document}

\maketitle
\begin{abstract}

We present \textbf{Disco4D}, a novel Gaussian Splatting framework for 4D human generation and animation from a single image. Different from existing methods, Disco4D distinctively disentangles clothings (with Gaussian models) from the human body (with SMPL-X model), significantly enhancing the generation details and flexibility. It has the following technical innovations. 
\textbf{1)} Disco4D learns to efficiently fit the clothing Gaussians over the SMPL-X Gaussians. \textbf{2)} It adopts diffusion models to enhance the 3D generation process, \textit{e.g.}, modeling occluded parts not visible in the input image. \textbf{3)} It learns an identity encoding for each clothing Gaussian to facilitate the separation and extraction of clothing assets. 
Furthermore, Disco4D naturally supports 4D human animation with vivid dynamics. 
Extensive experiments demonstrate the superiority of Disco4D on 4D human generation and animation tasks. 
Our visualizations can be found in \url{https://disco-4d.github.io/}.

 

\end{abstract}



 


\section{Introduction}
\begin{figure*}[tbhp]
    \centering
    \vspace{-15pt}
    \includegraphics[width=0.98\textwidth ,keepaspectratio]{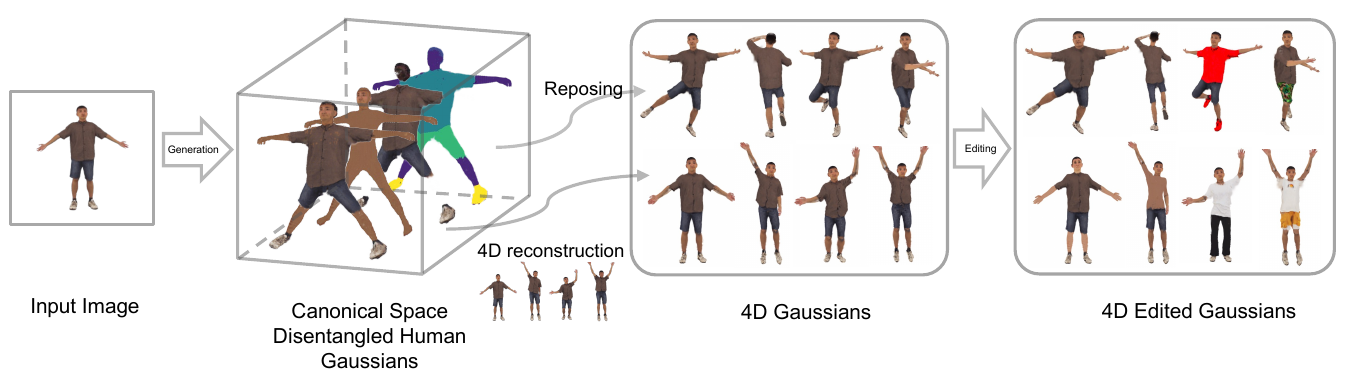}
    \vspace{-10pt}
    \caption{\small \textbf{Disco4D is a novel Gaussian Splatting framework for 4D disentangled human generation, animation and editing from a single image.}}
    \vspace{-10pt}    
    \label{figure:teaser}
\end{figure*}

The development of high-fidelity 3D digital humans is increasingly important across a variety of augmented and virtual reality applications. To streamline the creation of these digital avatars from easily accessible in-the-wild images, a multitude of research efforts have been made on reconstructing 3D clothed human models from a single image \cite{Alldieck2022,saito2019pifu,saito2020pifuhd,He2021,Huang2020,zheng2020pamir,xiu2023econ,xiu2022icon,xiu2023econ,hu2023sherf,huang2022elicit}. 
These works predominantly focus on the simultaneous reconstruction of the human body and clothing. 
Unfortunately, integrating them into applications that require virtual try-on or avatar customization poses significant challenges. This is primarily because the models are rendered as single-layer, non-animatable meshes where distinct attributes (e.g., hair, clothing, accessories) are merged into one continuous surface, with underlying layers completely obscured and self-contact areas inseparably connected. Such limitation complicates the re-animation and dynamic customization tasks. 

To address these issues, we propose Disco4D, a novel 4D clothed human reconstruction method that \textit{distinctly separates the human body from clothing elements}. It supports human animation as the 4th dimension, which cannot be realized by prior static 3D reconstruction works \cite{Alldieck2022,saito2019pifu,saito2020pifuhd, He2021,Huang2020,zheng2020pamir,xiu2023econ,xiu2022icon}. To achieve this, it employs the SMPL-X \cite{SMPLX2019} parametric model to represent the human body, capitalizing on its efficacy in capturing body structure and kinematics. Conversely, clothing, along with dynamic and variable elements such as hair and accessories, is represented using Gaussian models, which are able to model the large variability in clothing. 
By binding Gaussians to a SMPL-X model and fixing it during the training phase, Disco4D ensures the integrity of the body while focusing the learning process on the appearance aspects. 
\newhuien{To model occluded portions not visible in the input image, diffusion models are used to enhance the 3D generation process.}
Moreover, Disco4D includes an identity grouping mechanism for the Gaussians, which is instrumental in maintaining the separability and individuality of each clothing asset. 

The independent reconstruction of clothing and body offers several advantages. (1) \textit{Enhanced reconstruction fidelity}. The SMPL-X body serves as a stable anchor for the clothing to conform to. By isolating the focus to learn clothing gaussians, we achieve a more refined geometry and intricate detailing in the clothed model. (2) \textit{Fine-grained categorization and extraction of clothing items}. Disco4D is able to separate clothing Gaussians into their respective categories, which is crucial for the recovery and utilization of individual clothing assets. (3) \textit{Extensive editing capabilities}. Disco4D supports different editing functions, including the removal of specific items, inpainting (altering color or material), and other modifications. Such rich editing options allow for precise adjustments to individual assets without inadvertently affecting adjacent elements. This level of control is particularly beneficial in applications requiring detailed customization, such as virtual fashion design and digital content creation.
(4) \textit{Improved animation capabilities}. 
The body Gaussians adhere to the deformations dictated by the SMPL-X model, while clothing Gaussians conform to the underlying body movements but also exhibit behaviors true to their material characteristics. The disentangled deformation allows for nuanced adjustments to clothing behavior in response to complex body movements, thereby elevating the quality of clothed human animation.

\section{Related Works}

\subsection{3D Generation}
\begin{wraptable}{r}{0.5\textwidth}
\vspace{-15pt}
\caption{\textbf{Overview of 3D/4D generation methods from a single image.} }

\centering
\scriptsize
\setlength\tabcolsep{2pt}
\begin{tabular}{l|c c c}
\toprule
Method   &  Type  & Layered  & Animatable  \\
\hline
LGM~\cite{tang2024lgm} & General & \xmark & \xmark  \\

PiFU~\cite{saito2019pifu} & Human-centric & \xmark & \xmark \\
DreamFusion~\cite{poole2022dreamfusion} & General & \xmark & \xmark  \\

DreamGaussian~\cite{tang2023dreamgaussian} & General & \xmark & \xmark  \\

PiFU~\cite{saito2019pifu} & Human-centric & \xmark & \xmark \\

D-IF~\cite{yang2023dif} & Human-centric & \xmark & \xmark \\
HiLo~\cite{yang2024hilo} & Human-centric & \xmark & \xmark \\
ECON~\cite{xiu2023econ} & Human-centric & \xmark & \xmark \\

SHERF~\cite{hu2023sherf} & Human-centric & \xmark & \cmark \\



\hline
Disco4D & Human-centric & \cmark & \cmark \\

\bottomrule
\end{tabular}
\vspace{-7pt}

\label{table:overview_of_methods}
\end{wraptable}

\textbf{Single-image 3D Generation.}
%
Single-image reconstruction leverages advanced methods \cite{nichol2022pointe,jun2023shape} to generate 3D assets in the form of 3D point clouds or NeRF \cite{mildenhall2020nerf} from one image.
%
%
While earlier efforts using auto-encoders focused on synthetic objects \cite{Xu2019,chen2020bspnet,grf2020,duggal2022tars3D,szymanowicz2023splatter,shapenet2015}, newer approaches treat the task as conditional generation, employing diffusion models \cite{ho2020denoising} for 3D generation from both image and text \cite{ho2020denoising,melas2023realfusion,Tang_2023_ICCV,liu2023zero1to3,objaverseXL,rombach2021highresolution,poole2022dreamfusion,Magic123}. One-2-3-45 \cite{liu2023one2345} uses 2D diffusion models \cite{liu2023zero1to3,shi2023zero123plus}  to generate multi-view images for reconstruction, while LRM \cite{hong2023lrm} adopts transformer-based architecture to scale up the task on large datasets \cite{objaverseXL,yu2023mvimgnet}. Gaussian-based methods \cite{kerbl3Dgaussians}, particularly DreamGaussian \cite{tang2023dreamgaussian} and LGM \cite{tang2024lgm}, offer efficient, high-resolution 3D model generation from text or images.

\huien{Recently, video diffusion models have attracted significant attention due to their remarkable ability to generate intricate scenes and complex dynamics with great spatio-temporal consistency \cite{XuanyiLI2024Sora,blattmann2023stable,blattmann2023align,videoworldsimulators2024,guo2023animatediff, MakePixelsDance,bartal2024lumiere}. They are employed to generate consistent multi-view images, and then reconstruct underlying 3D assets with high quality \cite{chen2024v3d}.}

\noindent\textbf{Single-image human-centric 3D Generation.}
Significant research efforts have been made for 3D human reconstruction, which can be classified into the following categories. 
(1) \textit{Explicit-shape-based methods} rely on Human Mesh Recovery (HMR) using parametric models like SMPL \cite{SMPL2015} and SMPL-X \cite{SMPLX2019} to generate 3D body meshes \cite{Kanazawa2017,Kolotouros2019,Choi2020,Joo2021,Dwivedi2022,Kocabas2022,Kocabas2021,Feng2019,Choutas2020,Rong2021,Zhang2022,Moon2022}. To account for 3D garments, several approaches incorporate offsets \cite{zhu2019detailed,Xiang2020} or templates, utilize deformable garment templates \cite{jiang2020bcnet,bhatnagar2019mgn}, or employ non-parametric forms for clothed figures \cite{gabeur2019moulding,xiu2023econ,Zakharkin_2021_ICCV}. Despite their advancements, they face limitations in handling complex outfit variations and loose clothing due to inherent topological constraints.
(2) \textit{Implicit-function-based methods} utilize implicit representations like occupancy or distance fields for modeling clothed humans with complex geometries, such as loose garments. Techniques range from end-to-end regression of free-form implicit surfaces \cite{Alldieck2022,saito2019pifu,saito2020pifuhd} to use of geometric priors \cite{He2021,Huang2020,zheng2020pamir,xiu2023econ,xiu2022icon} and implicit shape completion \cite{xiu2023econ}. Notable works such as PIFu \cite{saito2019pifu}, ARCH(++) \cite{He2021,Huang2020}, and PaMIR \cite{zheng2020pamir} can extract textured models from images, but struggle with depth ambiguities and texture inconsistencies.
(3) \textit{NeRF-based methods} incorporate model-based priors (i.e., SMPL-X) for accurate human reconstruction, with efforts like SHERF \cite{hu2023sherf} and ELICIT \cite{huang2022elicit} improving reconstruction coherence by addressing 2D observation incompleteness leveraging appearance priors.
Most of these 3D clothed human reconstruction and animation works \cite{Alldieck2022,saito2019pifu,saito2020pifuhd, He2021,Huang2020,zheng2020pamir,xiu2023econ,xiu2022icon} require training on human-specific datasets, which brings another limitation on the availability of such datasets.

\noindent\textbf{3D Clothing Modeling.}
Reconstructing clothing from images and videos as a separate layer over the human body poses significant challenges due to the diversity of clothing topologies. Previous efforts relied on either template meshes or implicit surface models, and required extensive, high-quality 3D data from simulations \cite{bertiche2020cloth3d,patel20tailornet,santesteban2019virtualtryon,vidaurre2020virtualtryon} or tailored template meshes \cite{chen2021tightcap,halimi2022garment,Pons-Moll:Siggraph2017,Xiang_2021}. New methods were developed \cite{jiang2020bcnet,zhu2020deep} for multi-clothing models and versatile template meshes, respectively, facilitating diverse clothing topology encoding. However, these techniques typically fall short in capturing the clothing texture and appearance. The reliance on predefined clothing style templates further constrains their ability to handle real-world clothing variations. Corona et al. \cite{corona2021smplicit}  addressed these shortcomings by representing clothing layers with deep unsigned distance functions and an auto-decoder for style and cut differentiation, though this often produces overly-smooth reconstructions \cite{corona2021smplicit}. On the other hand, SCARF \cite{Feng2022scarf} and DELTA \cite{Feng2023DELTA} significantly enhance the visual fidelity by applying NeRF to clothing layers, but require self-rotating video inputs and considerable processing times.

\subsection{4D Animation}

\noindent\textbf{4D Animation.}
This task aims at capturing dynamic 3D scenes over time. Two primary approaches have emerged: modeling 4D scenes by adding time dimension $t$ or latent codes to spatial coordinates \cite{xian2021space,Gao-ICCV-DynNeRF,li2020neural}; combining deformation fields with static 3D scenes \cite{pumarola2020d,li2023dynibar,park2021nerfies,park2021hypernerf,du2021nerflow,tretschk2021nonrigid,yuan2021star}. Recent efforts in explicit or hybrid representations, like planar decomposition \cite{Cao2023HexPlane,kplanes_2023,shao2023tensor4d}, hash representations \cite{turki2023suds}, and other innovative methods \cite{Fang_2022,abou2022particlenerf,guan2022neurofluid}, have improved reconstruction speed and quality. Gaussian Splatting, especially, stands out for balancing efficiency with quality, with dynamic 3D Gaussians \cite{luiten2023dynamic} and 4D Gaussian Splatting \cite{wu20234dgaussians,yang2023deformable3dgs} techniques introducing time-dependent deformations to enhance reconstructions. Notably, DreamGaussian4D \cite{ren2023dreamgaussian4d} stands out by significantly reducing the optimization time while delivering high-quality 4D reconstructions.

 \begin{wraptable}{r}{0.42\textwidth}
\vspace{-15pt}
\caption{\textbf{Overview of 4D generation methods from video.} }

\centering
\scriptsize
\setlength\tabcolsep{2pt}
\begin{tabular}{l|c c}
\toprule
Method   &  Model unseen views  & Layered   \\
\hline
MonoHuman~\cite{yu2023monohuman} & \xmark & \xmark  \\
3DGS-Avatar~\cite{qian20233dgsavatar} &  \xmark & \xmark \\
Gaussian-Avatar~\cite{hu2024gaussianavatar} &  \xmark & \xmark  \\
GART~\cite{lei2023gart} & \xmark & \xmark  \\
Disco4D &  \cmark & \cmark \\

\bottomrule
\end{tabular}
\vspace{-7pt}

\label{table:overview_of_video_methods}
\end{wraptable}

\noindent\textbf{Human-centric 4D Animation.}
 \newhuien{Recent works leverage Gaussian-based methods \cite{hugs2024,li2024gaussianbody,liu2023animatable, yu2023monohuman, qian20233dgsavatar, hu2024gaussianavatar, lei2023gart} for 4D human reconstruction, requiring extensive frame sequences (50-100 frames) and/or multiple viewpoints. Currently there has not been any work on 4D layered human generation and animation from a single or few images, which will be achieved in this paper.}

\section{Methodology}
\label{section:methodology}



\subsection{Preliminary}
\label{section:preliminary}

\textbf{3D Gaussian Splatting} employs explicit 3D Gaussian points as its primary rendering entities. A 3D Gaussian point is defined as a function 
$
G(x) = e^{-\frac{1}{2}(x-\mu)^T \Sigma^{-1} (x-\mu)},
$
where $\mu$ and $\Sigma$ are the spatial mean and covariance matrix, respectively. Each Gaussian is also associated with its own rotation $r$, scaling $s$, opacity $\alpha$, a view-dependent color $c$ represented by spherical harmonic coefficients $f$. 

\textbf{SMPL-X parameterization}  \cite{SMPLX2019} is an extension of the SMPL body model \cite{SMPL2015} with face and hand, designed to capture a more accurate representation of intricate body movements. \textbf{SMPL-X} is defined as a function $M(\beta, \theta, \psi) : \mathbb{R}^{|\beta| \times |\theta| \times |\psi|} \rightarrow \mathbb{R}^{3N}$, parametrized by the pose $\theta \in \mathbb{R}^{3J}$ (where $J$ denotes the number of body joints), face and hands shape $\beta \in \mathbb{R}^{|\beta|}$ and facial expression $\psi \in \mathbb{R}^{|\psi|}$.

\subsection{Overview}
\label{subsection:DCH_generation}


\begin{figure*}[t]
    \centering
    \includegraphics[width=\linewidth ,keepaspectratio]{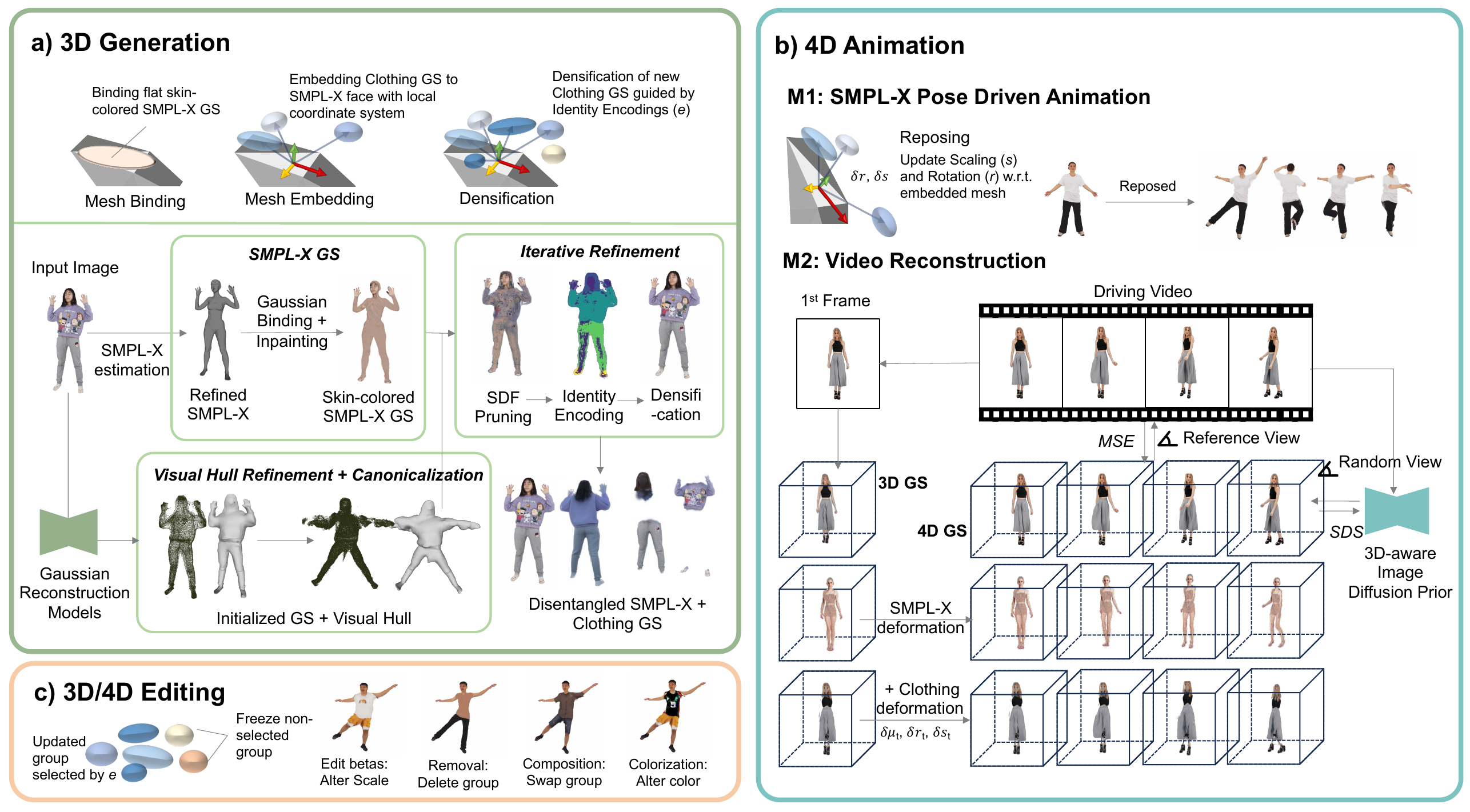}
    \vspace{-15pt}
    \caption{\small \textbf{Overview of Disco4D.} \textbf{(a) 3D Generation} utilizes a single image to obtain disentangled body and clothing Gaussians. Body, face and hand poses are refined to be pixel-aligned. For faster initialization, clothing Gaussians and visual hull are obtained with Gaussian Reconstruction Models. These clothing Gaussians are embedded to SMPL-X mesh and adopt the local coordinate system of the triangle. Subsequently, the iterative optimization process (pruning, identity encoding and densifying) separates the body and garments. The learned identity encodings guide the densification of the clothing Gaussians. \textbf{(b) 4D Animation} are achieved by either direct driving of SMPL-X poses or leveraging video to learn extra clothing deformation. Given a driving video, we first obtain a static 3D Disentangled GS model. Body and clothing Gaussians are deformed by pose transformations. We then optimize a deformation network to learn extra deformations for clothing GS at different timestamps. Various \textbf{(c) 3D/4D Editing} operations can be performed with our disentangled representation.
    } 
    \vspace{-15pt}
    \label{figure:architecture}
\end{figure*}





Given a single image, Disco4D generates animatable 3D clothed human avatars in a bottom-up manner, facilitating natural separability.
Our generated 3D clothed avatars, denoted as $S_{human}$, are represented as the concatenation of $S_{body}$ and $S_{cloth}$. Inspired by prior works~\cite{tang2023dreamgaussian,tang2024lgm}, $S$ capitalizes on Gaussian representations:
\begin{equation}
S = G(\mu, r, s, \alpha, c, e),
\end{equation}
where $\mu$, $r$, $s$, $\alpha$, $c$ and $e$ denote \emph{positions}, \emph{rotation}, \emph{scaling}, \emph{opacity}, \emph{spherical harmonics coefficients} and \emph{identity encoding}, respectively.
Different from traditional Gaussian representations, we add identity encoding $e$ to associate each Gaussian with its clothing category.

Figure \ref{figure:architecture} depicts our framework. The initial step is to generate colored SMPL-X Gaussians that represent the body beneath clothing (Sec.~\ref{subsection:smplx_gaussians}). 
Then we obtain a visual hull for canonicalization and refine the Gaussians predictions from \huien{Gaussian Reconstruction Models} to ensure they are properly aligned with and envelop the SMPL-X mesh (Sec.~\ref{subsection:hull_refinement}).
Next we iteratively optimize canonical clothing Gaussians external to the SMPL-X mesh (Sec.~\ref{subsection:optimization}). 
\lei{Lastly, we showcase the animation and editing of generated clothed avatars (Sec.~\ref{subsection:4d_animation}).
Notably, we leverage diffusion models to refine textures during 3D generation (Sec.~\ref{subsection:optimization}) and extrapolate unseen views during 4D animation (Sec.~\ref{subsection:4d_animation}).}



\subsection{SMPL-X Gaussians}
\label{subsection:smplx_gaussians}

Given an image, we first estimate coarse SMPL-X parameters with an off-the-shelf model~\cite{cai2024smpler}, and then refine coarse predictions by fitting on 2D keypoints and clothing segmentation masks~\cite{pavlakos2019expressive}, obtaining pixel-aligned SMPL-X parameters ($\beta, \theta, \psi$).

\noindent\textbf{Mesh Binding.} To convert the SMPL-X~\cite{SMPLX2019} mesh representation $M(\beta, \theta, \psi)$ into Gaussians $S_{body}$ for rendering, flat 3D Gaussians are bound directly on each mesh triangle face similar to SuGaR~\cite{guedon2023sugar}. The Gaussians means $\mu_{body}$ are explicitly computed using predefined barycentric coordinates in the corresponding triangles while the rotations of the Gaussians $r_{body}$ are derived from the surface normals. Initial scaling of the Gaussians $s_{body}$ is carefully selected to guarantee dense coverage of the mesh surface, thereby preventing any gaps. A specific scaling factor for the last axis is set to 0.1 to achieve a thin flat surface throughout.
In addition to structural representation, color representation beneath clothing is also a consideration. This is achieved by setting the opacity value $\alpha_{body}$ to 1.0 and optimizing a set of spherical harmonics $c_{body}$ for each Gaussian. The skin color in visible regions are supervised, while the skin color of occluded body Gaussians are encouraged to be similar to non-occluded regions. We minimize the difference between the body colors in the occluded region and the average skin color of the visible regions. \huien{$e_{body}$ is set to a fixed categorical label for rendering and is not updated during training.}
%
Notably, during the optimization of clothing Gaussians $S_{cloth}$, the parameters of SMPL-X Gaussians $S_{body}$ remain fixed. This preserves the integrity of the underlying body representation while allowing for flexible learning of clothing Gaussians.



\vspace{-5pt}
\subsection{Initialization of Clothing Gaussians}
\label{subsection:hull_refinement}
\label{subsection:clothing_gaussians}



Cloth styles are diverse, making proper initialization crucial for effective clothing modeling.
In synchronization with estimating SMPL-X, we first employ the Video Diffusion Model~\cite{blattmann2023stable} to estimate multi-view images.
Subsequently, we leverage Gaussian Reconstruction Models~\cite{tang2024lgm} to obtain initial 3D Gaussians and their corresponding visual hull.
Yet, the reconstructed 3D outputs often suffer from geometric inaccuracies, such as incorrect poses due to pose ambiguity or missing limbs.
To address this, we refine the coarse visual hull to ensure it accurately aligns with and overlays the SMPL-X mesh and encapsulates a good geometry for the clothed figure.
With SMPL-X aligned visual hull, we derive the refined Gaussians by adopting properties from their nearest neighbors.
The refined visual hull and Gaussians are then canonicalized for the optimization phase. 



\textbf{Mesh embedding.}
\label{subsection:mesh_embedding}
Each 3D clothing Gaussian is embedded on one triangle of the canonical mesh. The embedding directly defines the position of the Gaussians in both the canonical and posed space.
We set the mean position of the vertices as the origin of the local coordinate system. Given the vertices of a triangle, we define the mean position \( \mathbf{O} \) of the vertices as the origin of the local coordinate system. We define the position of a Gaussian $\mathbf{\mu} = \mathbf{O} + \mathbf{v}$ by an offset vector \( \mathbf{v} \) from the local origin. The displacement vector \( \mathbf{v} \) in local coordinates is given by $
\mathbf{v} = \sigma \mathbf{i} + \beta \mathbf{j} + \gamma \mathbf{k}$,
where \( \sigma \), \( \beta \), and \( \gamma \) are the components of the displacement vector along the \( \mathbf{i} \) (tangent), \( \mathbf{j} \) (bitangent), and \( \mathbf{k} \) (normal) respectively.
Unlike SplattingAvatar \cite{SplattingAvatar:CVPR2024}, which uses a displacement \( \mathbf{d} \) along the interpolated normal vector to position a Gaussian within the triangle plane, our method allows for embedding Gaussians in the non-nearest but most suitable triangle. For instance, hair Gaussians are tagged to the faces in the head instead of the nearest face used in other reposing methods~\cite{SplattingAvatar:CVPR2024,hu2023gauhuman}.
%



When the mesh is deformed for animation, the embedding provides additional rotation ($\delta r$) and scaling ($\delta s$) upon each Gaussian. Specifically, the Gaussian adopts the rotation of the embedding triangle face while its scaling is based on the change in lengths of the embedded triangles, not just the area. The scaling adjustments for each Gaussian are dynamically computed based on individual face transformations, providing finer control over the deformation. During optimization, both the Gaussian parameters and embedding parameters are updated simultaneously.


\vspace{-5pt}
\subsection{Optimization of Separable Gaussians}
\label{subsection:optimization}

With the SMPL-X Gaussian and initialized clothing Gaussian, we aim to optimize canonical clothing Gaussians $S_{cloth}$ outside the SMPL-X mesh. 
This involves three steps:
\textbf{1)} we use Signed Distance Function (SDF) loss and pruning to discourage and remove Gaussians that reside within the body;
\textbf{2)} we introduce \emph{identity encoding} $e$ to attach a clothing label for each clothing Gaussian, by lifting multi-view 2D segmentations of the target object onto the 3D Gaussians;
and \textbf{3)} guided by $e_{body}$ and $e_{cloth}$, we selectively densify only the relevant clothing points while ignoring body points. 
Once the disentangled clothing is obtained, we use SDS loss to in-paint high-resolution texture from the reference image to individual clothing Gaussians, thereby enriching the details of unseen regions.

\noindent\textbf{SDF Loss and Pruning.}
In reality, the clothing is always external to the body. During refinement, we ensure that the clothing Gaussians are positioned externally to the SMPL-X mesh by applying the SDF loss and a pruning strategy.
Specifically, the SDF loss $\mathcal{L}_{sdf}$ penalizes any new densified Gaussians that intrude into the space of the SMPL-X mesh, ensuring that the clothing Gaussians consistently remain outside the body's surface. Pruning is applied at fixed intervals to reinforce this separation, and systematically remove any Gaussians located within the SDF of the SMPL-X mesh.

\textbf{Identity encoding.} To associate each Gaussian to its clothing category, we introduce \emph{Identity Encoding ($e$)} as a new parameter to each Gaussian.
We adopt the clothing segmentation masks from SegFormer~\cite{Xie2021Segformer} as supervision, with labels for different categories. $e_{i}$ for each Gaussian is a learnable and compact vector of length 15, representing the remapped categories from the segmentation masks\footnote{Categories:  0: "Background", 1: "Hat", 2: "Hair", 3: "Sunglasses", 4: "Upper-clothes", 5: "Skirt", 6: "Pants", 7: "Dress",  8: "Belt", 9: "Left-shoe", 10: "Right-shoe", 11: "Face", 12: "Skin", 13: "Bag", 14: "Scarf"}.
During training, similar to Spherical Harmonic (SH) coefficients on representing the color of each Gaussian, we 
render these encoded identity vectors into 2D images in a differentiable manner following \cite{Ye2023gaussian_grouping}.
Given the rendered 2D features $E_{\text{id}}$, we take $softmax(E_{id})$ for identity classification, where $K$ is the total number of categories. We adopt a standard cross-entropy loss $\mathcal{L}_{2d}$ for ($K$+1)-category classification and an unsupervised 3D regularization loss $\mathcal{L}_{3d}$ to enforce spatial consistency and proximity among the top $k$-nearest 3D Gaussians' Identity Encodings.
Consequently, the overall identity loss is $\mathcal{L}_{id} = \mathcal{L}_{2d} + \mathcal{L}_{3d}$.
Refer to Appendix~\ref{subsection:id_implementation} for more details.

\noindent\textbf{Densification of clothing Gaussians.}
\newhuien{To learn clothing more efficiently}, we perform sampling for categorical Gaussians that belong to the same clothing category and embedding. We find the $k$-nearest Gaussian points for the resampled points and inherit their Gaussian properties (scaling, rotation, opacity, SH properties).
By selectively densifying clothing Gaussians, we only add necessary Gaussians while ignoring body Gaussians.

\noindent\textbf{Anisotropy.} To prevent overly-skinny kernels that point outward from the object surface under large deformations, we enforce the anisotropy of Gaussian kernels following \cite{xie2023physgaussian}.
During optimization, we employ $ \mathcal{L}_{\text{ani}} = \frac{1}{|P|} \sum_{p \in P} \max\left(\frac{\max(s_p)}{\min(s_p)}, \tau\right){-\tau} $ where \huien{$s_p$} is the scalings of 3D Gaussians. This loss essentially constrains that the ratio between the major and minor axis lengths does not exceed \huien{$\tau$}.



\noindent\textbf{Total loss. }
To inpaint occluded textures, we use the $\mathcal{L}_{SDS}$ loss on the Gaussians in the canonical pose. This step follows the optimization of the front view for 500 training steps.
Combined with the conventional 3D Gaussian Loss $L_{ori}$ on image rendering, the total loss $L$ for end-to-end training is:
\vspace{-5pt}
\begin{equation}
\mathcal{L} = \mathcal{L}_{ori}  + \mathcal{L}_{id} + \mathcal{L}_{ani} + \mathcal{L}_{sdf} + \mathcal{L}_{SDS} 
\end{equation}
\vspace{-5pt}

\vspace{-10pt}
\subsection{4D Human Animation and Editing}
\label{subsection:4d_animation}
\vspace{-10pt}




Disco4D's disentangled representation naturally supports animation and editing.
The canonical Gaussians $S_{body}$ and $S_{cloth}$ enable separate deformations for clothing and body, ensuring realistic animation.
Besides, individual clothing categories can be easily edited using image or text prompts. The learned clothing can be transferred to different body shapes and poses, for versatile customization.

\noindent\textbf{Animating Gaussians.} 
\huien{
As shown in Figure \ref{figure:architecture}, Disco4D enables animation of the canonical human Gaussian via two methods.
Firstly, Gaussians can be directly driven using 3D SMPL-X sequences obtained from a motion database or estimated from 2D videos. 
Secondly, Disco4D enhances the model by learning detailed clothing dynamics from monocular videos. This disentanglement enables the focused modeling of clothing dynamics without altering the underlying human representation.
}

To extend static 3D Gaussians into dynamic 4D Gaussians, a deformation network is trained to predict changes in position, rotation, and scale of the reposed clothing Gaussians based on a timestamp, as described in DreamGaussian4D~\cite{ren2023dreamgaussian4d}.
Unlike \cite{ren2023dreamgaussian4d}, which learns deformations for all Gaussians, Disco4D models body Gaussians using the SMPL-X mesh, while clothing Gaussians employ posed transformations and learned deformations.
The transformation is defined as $S'' =   \phi(S', t)$ where $\phi$ is the deformation network,  $S'$ is the spatial descriptions of the reposed 3D clothing Gaussian, $t$ is the timestamp, and $S''$ is the spatial descriptions of the deformed and reposed 3D clothing Gaussians.
Following \cite{ren2023dreamgaussian4d}, the deformation model is initialized to predict zero deformation at the start of training to avoid divergence between dynamic and static models. The weights and biases of the final prediction heads are initialized to zero, and skip connections are introduced to enable gradient backpropagation.


To optimize the deformation field using the reference view video, we minimize the reconstruction loss $\mathcal{L}_{Ref}$ between the rendered image and video frame at each timestep. 
To propagate the motion from the reference view to the entire 3D model, we leverage Zero-1-to-3-XL \cite{objaverseXL} to predict the deformation of the unseen part to calculate $\mathcal{L}_{SDS}$. 
Despite per-frame predictions of image diffusion models, the fixed color and opacity of static 3D Gaussians help preserve temporal consistency.






\noindent\textbf{Editing Clothing Gaussians.}
We extract the Gaussians corresponding to the specific category and edit them. This allows fine-grained editing and ensures that other Gaussians are not affected. Instead of fine-tuning all 3D Gaussians, we freeze the properties for most of the well-trained Gaussians and only adjust a small part of 3D Gaussians relevant to the target categories. For 3D object removal, we simply delete the 3D Gaussians of the editing target. For 3D object colorization by in-painting or text guidance, we reinitialise the color and tune the color (SH) parameters of the corresponding Gaussian group, while fixing the 3D positions and other properties to preserve the learned 3D geometry. 








\section{Experiments}
\vspace{-5pt}
Our detailed implementation and experiment setup can be found in Appendix \ref{subsection:training_details}.






\subsection{3D Generation}
\vspace{-8pt}  
\begin{table}[!htbp]
  \caption{\small\textbf{CLIP-embedding loss for generated humans and segmented assets, and performance (PSNR, SSIM, LPIPS) comparisons for novel poses and views on the Synbody and CloSe datasets across DreamGaussian, LGM, SHERF, and Disco4D.}}
  \label{table:3d_recon_comparison}
  \centering
  \tiny

  \begin{adjustbox}{max width=\linewidth}
  \begin{tabular}{lcccccccccccccccccccccccc}
    \toprule
    \multirow{2}{*}{\textbf{Method}} & \multicolumn{7}{c}{\textbf{SynBody}} & \multicolumn{10}{c}{\textbf{CloSe}}  \\
        \cmidrule(lr){2-8} \cmidrule(lr){9-18} 
     & \multicolumn{4}{c}{\textbf{CLIP}} & \multicolumn{3}{c}{\textbf{Novel View}} & \multicolumn{4}{c}{\textbf{CLIP}} & \multicolumn{3}{c}{\textbf{Novel View}} & \multicolumn{3}{c}{\textbf{Novel Pose}}\\
    \cmidrule(lr){2-5} \cmidrule(lr){6-8} \cmidrule(lr){9-12} \cmidrule(lr){13-15} \cmidrule(lr){16-18} 
     & All $\uparrow$    &   Pants $\uparrow$ &     Shirt $\uparrow$ &    Shoes $\uparrow$ & PSNR $\uparrow$ & SSIM $\uparrow$ & LPIPS $\downarrow$ & All $\uparrow$    &   Pants $\uparrow$ &     Shirt $\uparrow$ &    Shoes $\uparrow$ & PSNR $\uparrow$ & SSIM $\uparrow$ & LPIPS $\downarrow$ & PSNR $\uparrow$  & SSIM $\uparrow$ & LPIPS $\downarrow$\\
    
    \midrule
    DreamGaussian~\cite{tang2023dreamgaussian} & 0.751 & 0.715 & 0.710 & 0.749 & 13.118 & 0.883 & 0.229  & 0.734 & 0.693 & 0.674 & 0.767 & 20.08 & 0.939 & 0.089 & - & - & - \\
    LGM~\cite{tang2024lgm} &     0.807 & 0.724 & 0.747 & 0.760 & 12.884 &  \textbf{0.876} & 0.228 & 0.829 & 0.727 & 0.712 & 0.778 & \textbf{20.50}  & \textbf{0.939} & \textbf{0.077} & - & - & - \\
    SHERF~\cite{hu2023sherf} & 0.766 & 0.649  & 0.636 & 0.714 & 15.189 & 0.852 & 0.189 & 0.777 & 0.785 & 0.729 & 0.801 & 18.96 & 0.912 & 0.083 & 15.54 & 0.844 & 0.165 \\
    Ours  &      \textbf{0.851} &     \textbf{0.784} &      \textbf{0.753}  &   \textbf{0.801}  &  \textbf{15.691} &    0.848 &      \textbf{0.185}   & \textbf{0.856} &    \textbf{0.858} &     \textbf{0.810}  &  \textbf{0.842}  & 20.10 & 0.918 & 0.081 & \textbf{17.96} & \textbf{0.851} & \textbf{0.136}\\
    \bottomrule
    
  \end{tabular}
  \end{adjustbox}
\vspace{-10pt}  
\end{table}
\begin{figure*}[!htbp]
    \centering
    \includegraphics[width=0.98\textwidth ,keepaspectratio]{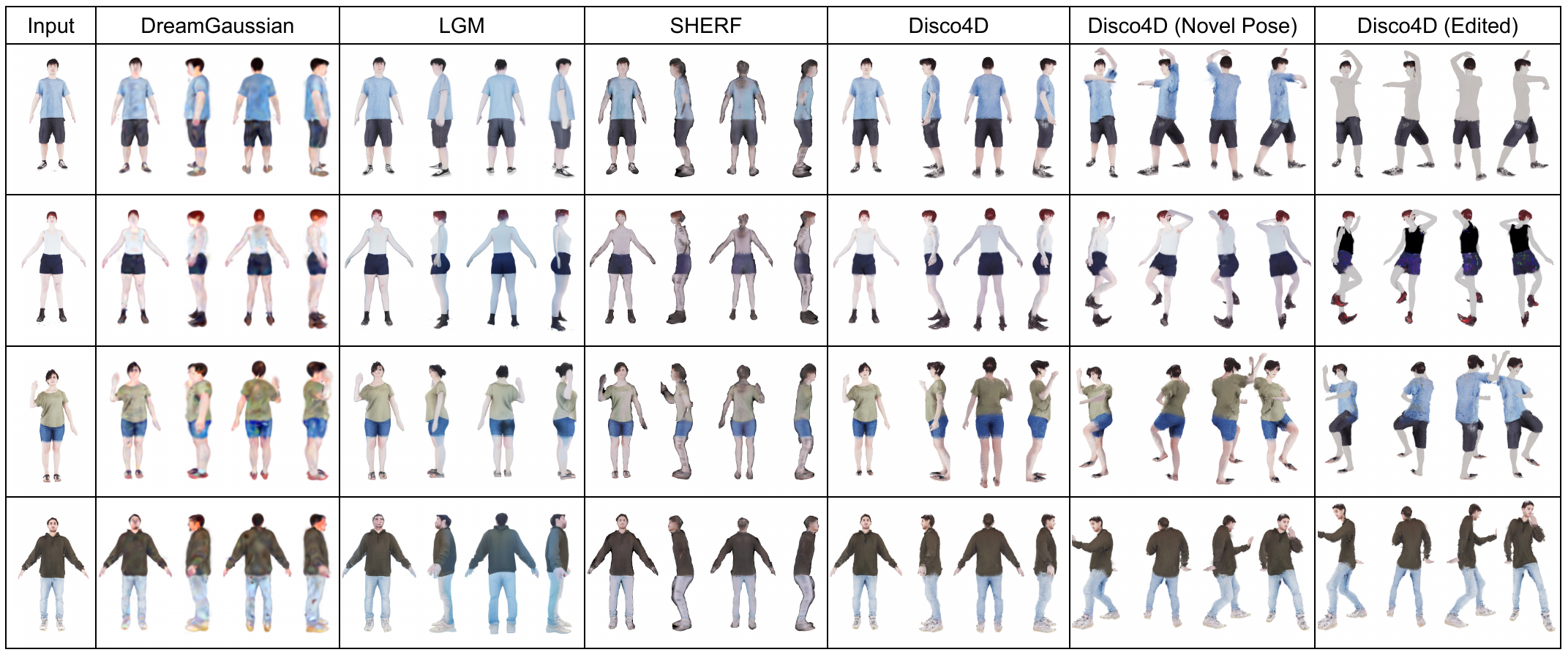}
    \caption{\small \textbf{Qualitative comparison of Image generation across DreamGaussian, LGM, SHERF, and Disco4D.}}
    \vspace{-5pt}    
    \label{figure:qualitative_close}
\end{figure*}

\noindent\textbf{Generation and Disentanglement.} Our generation and disentanglement results are presented in Figure \ref{figure:qualitative_close} and Table \ref{table:3d_recon_comparison}. We assessed the disentanglement quality using the Synbody \cite{yang2023synbody} and CloSe \cite{antic2024close} datasets, rendering 30 and 110 clothed human meshes respectively from four angles and evaluating CLIP-similarity, PSNR, SSIM, and LPIPS for various poses and views within the CloSe dataset. Our method leverages on diffusion models without the need for training on human specific datasets. Therefore, we compare it with DreamGaussian \cite{tang2023dreamgaussian} and LGM \cite{tang2024lgm} that reconstruct 3D objects from diffusion models. Additionally, we conducted comparisons with SHERF, a human-centric baseline for evaluating novel poses and views. We observe from Figure \ref{figure:qualitative_close} that our method has higher fidelity and better geometry for body parts such as face and limbs due to the representation using SMPL-X Gaussians. 
We outperform DreamGaussian, LGM and SHERF on benchmarks for SynBody \cite{yang2023synbody} and CloSe \cite{antic2024close}.


\noindent\textbf{Editing.} We can edit specific clothing appearance given an image or text prompt, repose the person and transfer person characteristics. The disentanglement allows fine-grained editing and modification of individual assets without affecting other assets, and stacking multiple edits (Figure \ref{figure:qualitative_close}).

\begin{wraptable}{r}
{0.5\textwidth} 
\vspace{-10pt}
  \caption{\small\textbf{
  User study rates quality of generated 3D Gaussians from 1-5, the higher the better.}}
  \label{table:user_study}
  \centering
  \scriptsize
  \vspace{-5pt}
  \setlength\tabcolsep{2pt}
  \begin{tabular}{@{}lcc@{}}
    \toprule
     Metric & Image Consistency $\uparrow$    &   Overall Quality $\uparrow$  \\
    \midrule
    DreamGaussian~\cite{tang2023dreamgaussian}  &     2.017  & 1.852 \\
    LGM~\cite{tang2024lgm}  &      2.338   &    2.017 \\
    Disco4D (Ours)  &    \textbf{3.142} &  \textbf{3.037} \\
    
  \bottomrule
  \end{tabular}
  \vspace{-8pt}
\end{wraptable}
\textbf{User study.}  We conducted a user study to evaluate the generative quality of our image-to-3D Gaussians reconstruction on random in-the-wild images from SHHQ, detailed in Table \ref{table:user_study}. This study focuses on reference view consistency and overall generation quality, crucial aspects in image reconstruction tasks. We rendered 360-degree rotation videos for 25 images generated by DreamGaussian, LGM, and our method. We invited 43 volunteers to rate 24$\sim$27 mixed samples from these methods on image consistency and overall model quality, yielding 1080 valid scores. As shown in Table \ref{table:user_study}, our method was preferred, demonstrating better alignment with the original image content and superior overall quality.





\subsection{4D animation}

\begin{wrapfigure}{r}
{0.6\textwidth} 
\vspace{-20pt}
    \centering
    \includegraphics[width=0.6\textwidth, keepaspectratio]{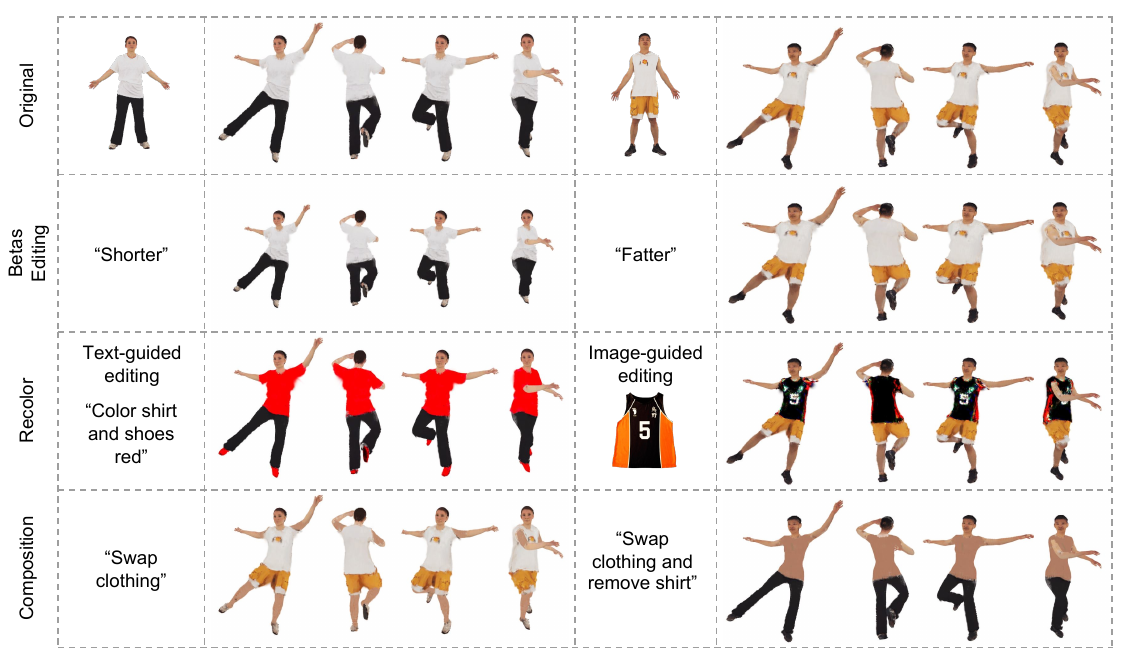}
    \caption{\small \textbf{First frame Editing and Animation}. Betas Editing, Recoloring (Text/Image-guided), Composition (Removal, Swap).}
    \vspace{-20pt}    
    \label{figure:4d_editing}
\end{wrapfigure}

\noindent\textbf{Pose-Driven Animation.} 
Our method generates canonical Gaussians that can be animated with any pose sequence. Figure \ref{figure:2d_animation_comparison} in the Appendix demonstrates our animation capabilities and compares them with current SOTA 2D animation methods. Using identical inputs—a single frame and pose sequence—our approach more effectively preserves the body shape and fine details such as facial features and clothing. It surpasses Animate-Anyone \cite{hu2023animateanyone} and Magic-Animate \cite{xu2023magicanimate} in accurately modeling fine-grained body parts like hands and faces, and exhibits greater consistency compared to CHAMP \cite{zhu2024champ}. Our method's disentanglement feature further allows for direct manipulation of Clothing Gaussians, as shown in Figure \ref{figure:4d_editing}.

\begin{table}[!htbp]
  \caption{\small\textbf{CLIP-embedding loss for generated humans and segmented assets, and performance (PSNR, SSIM, LPIPS) comparison on the 4D-Dress dataset across various video-to-4D methods. }}
  \label{table:ablation_4ddress}
  \centering
  \tiny
  \begin{tabular}{@{}llllll@{}}
    \toprule
     & All $\uparrow$    &   Assets $\uparrow$ & PSNR $\uparrow$    &   SSIM $\uparrow$  & LPIPS $\downarrow$   \\
    \midrule
    DreamGaussian4D~\cite{ren2023dreamgaussian4d}  &      0.784 & 0.769 & 20.54 & 0.93 & 0.080  \\
    MonoHuman~\cite{yu2023monohuman}  &      0.762 & 0.743 & 20.22 & 0.92 & 0.086  \\
    GART~\cite{lei2023gart}  &      0.800 & 0.772 & 18.81 & 0.92 & 0.086  \\
    GaussianAvatar~\cite{hu2024gaussianavatar}  &      0.822 & 0.768 & 20.01 & 0.93 & 0.069  \\
    DreamGaussian4D (LGM init)  &      0.809 &  0.795  & 19.16 & 0.93 & 0.086  \\
    DreamGaussian4D (Disco4D init)  &      0.870 &  0.849  & 21.02 & 0.93 & 0.065  \\
    Disco4D (reposed)  &      0.853 &  0.774  & 23.94 & 0.95 & 0.049   \\
    Disco4D (reposed) + learned deformations  &    \textbf{0.900} &  \textbf{0.865}  &  \textbf{25.46} &  \textbf{0.96} &  \textbf{0.035}  \\
  \bottomrule
    \end{tabular}
\end{table}

\noindent\textbf{4D Reconstruction.} For the 4D-Dress Dataset\cite{wang20244ddress}, we evaluated 8 sequences, assessing CLIP similarity scores against ground-truth meshes and disentangled assets, along with novel view performance (PSNR, SSIM, LPIPS) from four different viewpoints. The quantitative results of our generation approach are summarized in Table \ref{table:ablation_4ddress}, where we benchmark our method against existing video-to-4D general GS approaches, such as DreamGaussian4D \cite{ren2023dreamgaussian4d}, as well as human-centric GS methods, including MonoHuman \cite{yu2023monohuman}, GART \cite{lei2023gart}, and GaussianAvatar \cite{hu2024gaussianavatar}.

\begin{figure*}[!htbp]
    \centering
    \includegraphics[width=0.95\textwidth ,keepaspectratio]{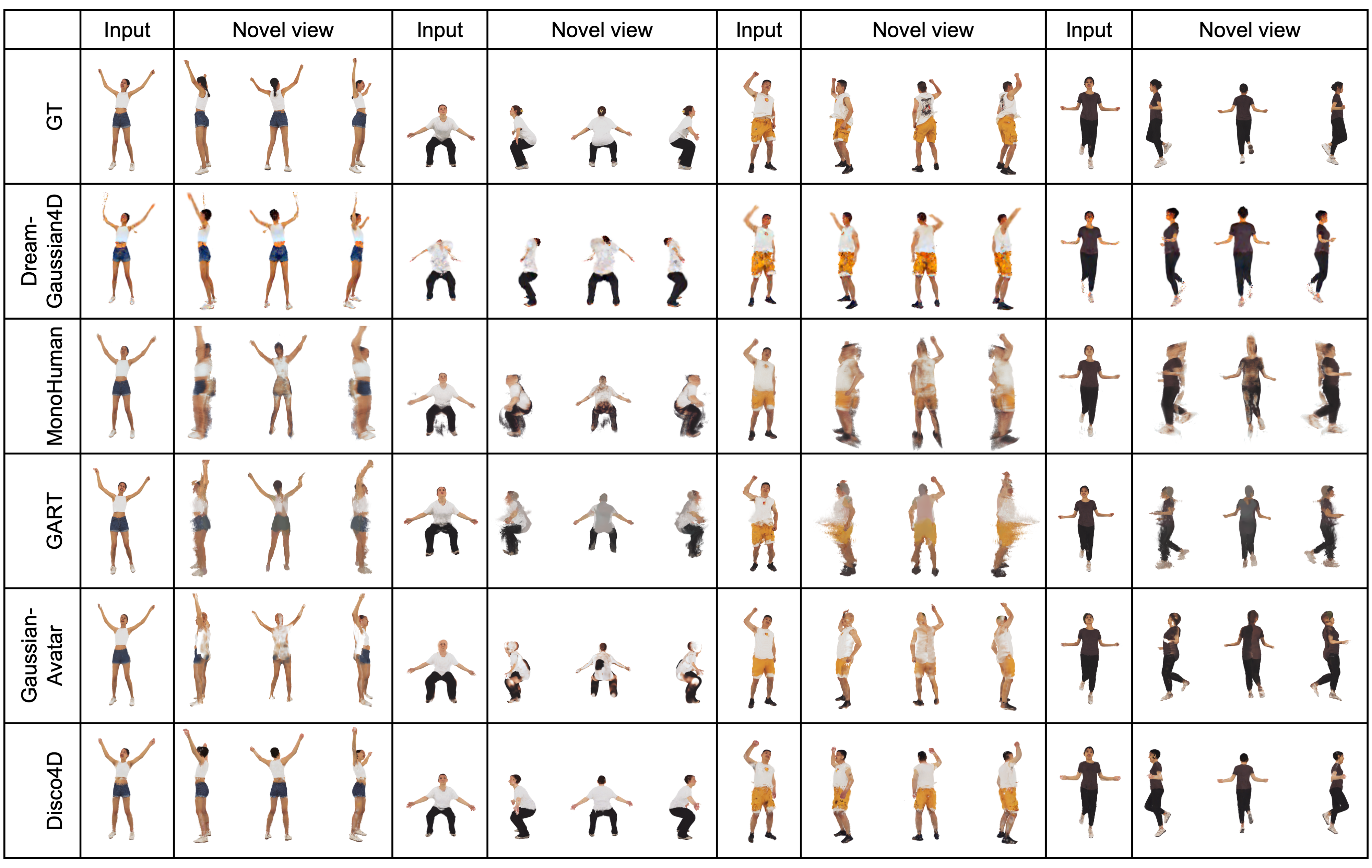}
    \caption{\small \textbf{Qualitative comparison of 4D generation between DreamGaussian4D, MonoHuman, GART, GaussianAvatar, and Disco4D.}} 
    \label{figure:4ddress_vid}
\end{figure*}

We outperform MonoHuman \cite{yu2023monohuman}, GART \cite{lei2023gart}, and GaussianAvatar \cite{hu2024gaussianavatar} (Table \ref{table:ablation_4ddress}) as these methods reconstruct using known video information, unable to model unseen regions. Consequently, these methods cannot accurately model back views from front-facing videos, leading to artifacts in other perspectives and canonical space (see Figure \ref{figure:4ddress_vid}). In contrast, our method first performs reconstruction and subsequently incorporates details, such as clothing deformation, from the input frames, enabling consistent reconstruction even in unseen viewpoints. 

\begin{figure*}[htbp]
    \centering
    \includegraphics[width=\textwidth ,keepaspectratio]{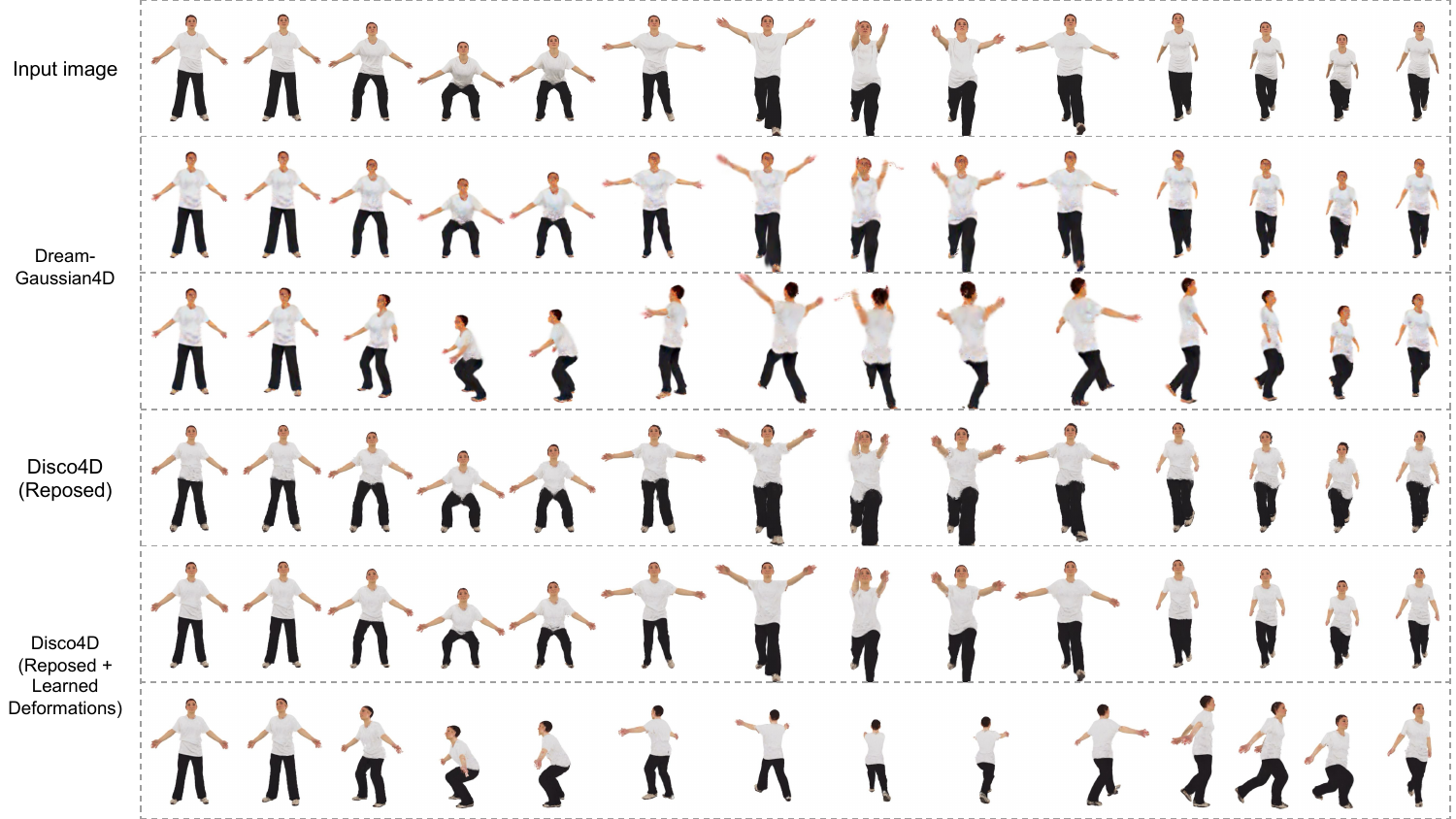}
    \caption{\small \textbf{4D reconstruction results on 4D-Dress Dataset.}}
    \label{figure:4d_reconstruction}
\end{figure*}

While DreamGaussian4D \cite{ren2023dreamgaussian4d} is capable of modeling back-view information, the details remain coarse. Our results demonstrate that initializing with our model from the first frame (DreamGaussian4D Disco4D-init) significantly outperforms other initialization methods (DreamGaussian4D-LGM init, DreamGaussian init) in both fidelity and geometry (Table \ref{table:ablation_4ddress}). Nevertheless, without incorporating human priors, DreamGaussian4D \cite{ren2023dreamgaussian4d} still faces challenges, such as missing limbs and difficulty modeling fine details like facial features (see Figure \ref{figure:4d_reconstruction}).

Reposing our canonical avatar enables us to align the body and assets accurately with the inferred postures from the source video, yielding high-quality reconstruction of faces, hands, and garments. Our reposed method surpasses DreamGaussian4D in geometry and fidelity by incorporating human priors. However, reposing alone cannot capture clothing dynamics. To address this, our disentangled approach models clothing deformations on the reposed Gaussians, guided by a diffusion model.  As demonstrated in Figure \ref{figure:4d_reconstruction} and Table \ref{table:ablation_4ddress}, this process enhances the accuracy of clothing resemblance to the ground truth. The combination of asset repositioning and learned deformations improves modeling quality, with repositioning handling pose-driven changes and learned deformations simulating dynamic asset movements as observed in the driving video.

\huien{
\noindent\textbf{4D Editing.} 
For a normal pipeline in character animation, editing the person in the video requires high consistency throughout all frames. For pose-driven animation methods, first frame editing and generation is required. Our method directly edits the gaussians, which is more straightforward, fine-grained and consistent. This is seen from Figure \ref{figure:4d_editing}.
}






\clearpage

\subsection{Ablation Studies}



 
\textbf{Initialization of clothing Gaussians.} This process is crucial for high fidelity reconstruction. As shown in Figure \ref{figure:ablation_initialisation} in the Appendix, we evaluate different strategies, including random, surface, and hull-based initialization. Hull-based initialization significantly enhances the model accuracy and realism over other methods. Initialization directly on the SMPL-X surface often leads to inaccurate geometries, particularly with complex or loose garments, creating elongated, thin Gaussians and visual artifacts. In contrast, hull-based initialization captures garment details more effectively and maintains pose consistency, closely aligning with the true geometry of the clothed body.

\begin{figure*}[htb!]
    \centering
    \vspace{-10pt}
    \includegraphics[width=\textwidth ,keepaspectratio]{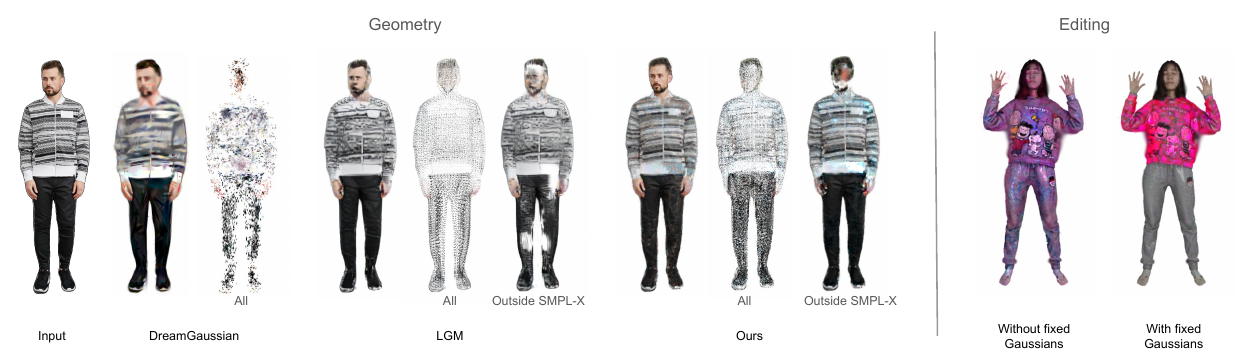}
    \vspace{-10pt}
    \caption{\small \textbf{Ablation of points geometry (left) and editing results (right). Points ("All") are visualised with a Gaussian Scale of 0.1. }}
    \label{figure:ablation}
    \vspace{-10pt}
\end{figure*}
\noindent\textbf{Geometry of Clothing Gaussians.} 
Figure \ref{figure:ablation} highlights the differences in clothing geometry between DreamGaussian \cite{tang2023dreamgaussian}, LGM \cite{tang2024lgm} and our method. 
In DreamGaussian, all points are confined within the body geometry, whereas in LGM, about half of the points extend beyond the SMPL-X body. Removing internal points leaves sparse, translucent representations for clothing. This sparsity suggests reliance on internal points for visual representation, failing to accurately depict the object's geometry where appearance should primarily originate from surface points. Often, clothing Gaussian points are incorrectly positioned inside the body's hull rather than on the surface. To better represent clothing geometry, our method positions all clothing Gaussians externally to the SMPL-X body mesh, accurately reflecting the garment's actual physical characteristics (Figure \ref{figure:ablation}).


\noindent\textbf{Clothing editing.} Figure \ref{figure:ablation} shows our editing results with the prompt "Color the top pink". Disco4D allows for precise editing of the targeted clothing without affecting other areas.

\section{Conclusion}
\label{section: conclusion}


We propose Disco4D, a novel approach for the generation of 3D animatable clothed human Gaussians from a single image, emphasizing high-fidelity detail and separation of assets. We manage to compositionally generate separate components, such as haircut, accessories, and decoupled outfits. Our core insight is the fixing of SMPL-X Gaussians, fitting segmented Gaussians over SMPL-X Gaussians, and application of diffusion models to enhance 3D reconstruction, including modeling occluded parts not visible in the input image. Its capability to separate assets offers significant advantages, including localized, fine-grained editing of individual assets and enhanced animatability. 


\noindent\textbf{Limitations and Future Works.}
Despite achieving impressive results, some failure cases still exist, as shown in Figure \ref{figure:failure_cases} in the Appendix. Disco4D relies on robust and pixel-aligned SMPL-X estimation, which is still an unsolved problem. Disco4D occasionally fails for poor visual hull initialization. 
The extraction of mesh assets from clothing Gaussians using Local Density Query, as per DreamGaussian \cite{tang2023dreamgaussian}, currently loses fine-grained details. Enhancing the detail level of geometry derived from clothing Gaussians could bolster the utility of reconstructed assets in animation and simulation applications. Furthermore, the initialized visual hulls obtained from multi-view SMPL-X guided images are often of suboptimal quality and suffer from poor side and back views, necessitating refinement. Improving pose guidance models to achieve more accurate visual hulls could alleviate the need for extensive refinement. In addition, future works could look into modeling multi-layered clothing and reconstructing the occluded clothing. Current animation only supports a small number of frames, future works could look into modelling long sequences. 
Disco4D has many positive applications. On the other hand, it has the potential to facilitate deepfake avatars and raise IP concerns. Regulations should be built to address these issues alongside its benefits in the entertainment industry.

\clearpage


\bibliography{neurips_2024}

\begin{thebibliography}{120}
\providecommand{\natexlab}[1]{#1}
\providecommand{\url}[1]{\texttt{#1}}
\expandafter\ifx\csname urlstyle\endcsname\relax
  \providecommand{\doi}[1]{doi: #1}\else
  \providecommand{\doi}{doi: \begingroup \urlstyle{rm}\Url}\fi

\bibitem[Abou-Chakra et~al.(2022)Abou-Chakra, Dayoub, and S{\"u}nderhauf]{abou2022particlenerf}
Jad Abou-Chakra, Feras Dayoub, and Niko S{\"u}nderhauf.
\newblock Particlenerf: Particle based encoding for online neural radiance fields.
\newblock \emph{arXiv preprint arXiv:2211.04041}, 2022.

\bibitem[Alldieck et~al.(2022)Alldieck, Zanfir, and Sminchisescu]{Alldieck2022}
Thiemo Alldieck, Mihai Zanfir, and Cristian Sminchisescu.
\newblock {Photorealistic Monocular 3D Reconstruction of Humans Wearing Clothing}.
\newblock \emph{Proceedings of the IEEE Computer Society Conference on Computer Vision and Pattern Recognition}, 2022-June:\penalty0 1496--1505, 2022.
\newblock ISSN 10636919.
\newblock \doi{10.1109/CVPR52688.2022.00156}.

\bibitem[Antić et~al.(2024)Antić, Tiwari, Ozcomlekci, Marin, and Pons-Moll]{antic2024close}
Dimitrije Antić, Garvita Tiwari, Batuhan Ozcomlekci, Riccardo Marin, and Gerard Pons-Moll.
\newblock {CloSe}: A {3D} clothing segmentation dataset and model.
\newblock In \emph{International Conference on 3D Vision (3DV)}, March 2024.

\bibitem[Bar-Tal et~al.(2024)Bar-Tal, Chefer, Tov, Herrmann, Paiss, Zada, Ephrat, Hur, Liu, Raj, Li, Rubinstein, Michaeli, Wang, Sun, Dekel, and Mosseri]{bartal2024lumiere}
Omer Bar-Tal, Hila Chefer, Omer Tov, Charles Herrmann, Roni Paiss, Shiran Zada, Ariel Ephrat, Junhwa Hur, Guanghui Liu, Amit Raj, Yuanzhen Li, Michael Rubinstein, Tomer Michaeli, Oliver Wang, Deqing Sun, Tali Dekel, and Inbar Mosseri.
\newblock Lumiere: A space-time diffusion model for video generation, 2024.

\bibitem[Bertiche et~al.(2020)Bertiche, Madadi, and Escalera]{bertiche2020cloth3d}
Hugo Bertiche, Meysam Madadi, and Sergio Escalera.
\newblock Cloth3d: clothed 3d humans.
\newblock In \emph{European Conference on Computer Vision}, pp.\  344--359. Springer, 2020.

\bibitem[Bhatnagar et~al.(2019)Bhatnagar, Tiwari, Theobalt, and Pons-Moll]{bhatnagar2019mgn}
Bharat~Lal Bhatnagar, Garvita Tiwari, Christian Theobalt, and Gerard Pons-Moll.
\newblock Multi-garment net: Learning to dress 3d people from images.
\newblock In \emph{{IEEE} International Conference on Computer Vision ({ICCV})}. {IEEE}, oct 2019.

\bibitem[Blattmann et~al.(2023{\natexlab{a}})Blattmann, Dockhorn, Kulal, Mendelevitch, Kilian, Lorenz, Levi, English, Voleti, Letts, Jampani, and Rombach]{blattmann2023stable}
Andreas Blattmann, Tim Dockhorn, Sumith Kulal, Daniel Mendelevitch, Maciej Kilian, Dominik Lorenz, Yam Levi, Zion English, Vikram Voleti, Adam Letts, Varun Jampani, and Robin Rombach.
\newblock Stable video diffusion: Scaling latent video diffusion models to large datasets, 2023{\natexlab{a}}.

\bibitem[Blattmann et~al.(2023{\natexlab{b}})Blattmann, Rombach, Ling, Dockhorn, Kim, Fidler, and Kreis]{blattmann2023align}
Andreas Blattmann, Robin Rombach, Huan Ling, Tim Dockhorn, Seung~Wook Kim, Sanja Fidler, and Karsten Kreis.
\newblock Align your latents: High-resolution video synthesis with latent diffusion models, 2023{\natexlab{b}}.

\bibitem[Brooks et~al.(2024)Brooks, Peebles, Holmes, DePue, Guo, Jing, Schnurr, Taylor, Luhman, Luhman, Ng, Wang, and Ramesh]{videoworldsimulators2024}
Tim Brooks, Bill Peebles, Connor Holmes, Will DePue, Yufei Guo, Li~Jing, David Schnurr, Joe Taylor, Troy Luhman, Eric Luhman, Clarence Ng, Ricky Wang, and Aditya Ramesh.
\newblock {Video generation models as world simulators}.
\newblock 2024.
\newblock URL \url{https://openai.com/research/video-generation-models-as-world-simulators}.

\bibitem[Cai et~al.(2024)Cai, Yin, Zeng, Wei, Sun, Yanjun, Pang, Mei, Zhang, Zhang, et~al.]{cai2024smpler}
Zhongang Cai, Wanqi Yin, Ailing Zeng, Chen Wei, Qingping Sun, Wang Yanjun, Hui~En Pang, Haiyi Mei, Mingyuan Zhang, Lei Zhang, et~al.
\newblock Smpler-x: Scaling up expressive human pose and shape estimation.
\newblock \emph{Advances in Neural Information Processing Systems}, 36, 2024.

\bibitem[Cao \& Johnson(2023)Cao and Johnson]{Cao2023HexPlane}
Ang Cao and Justin Johnson.
\newblock Hexplane: A fast representation for dynamic scenes.
\newblock \emph{CVPR}, 2023.

\bibitem[Chang et~al.(2015)Chang, Funkhouser, Guibas, Hanrahan, Huang, Li, Savarese, Savva, Song, Su, Xiao, Yi, and Yu]{shapenet2015}
Angel~X. Chang, Thomas Funkhouser, Leonidas Guibas, Pat Hanrahan, Qixing Huang, Zimo Li, Silvio Savarese, Manolis Savva, Shuran Song, Hao Su, Jianxiong Xiao, Li~Yi, and Fisher Yu.
\newblock {ShapeNet: An Information-Rich 3D Model Repository}.
\newblock Technical Report arXiv:1512.03012 [cs.GR], Stanford University --- Princeton University --- Toyota Technological Institute at Chicago, 2015.

\bibitem[Chen et~al.(2021)Chen, Pang, Wei, Peihao, Xu, and Yu]{chen2021tightcap}
Xin Chen, Anqi Pang, Yang Wei, Wang Peihao, Lan Xu, and Jingyi Yu.
\newblock Tightcap: 3d human shape capture with clothing tightness field.
\newblock \emph{ACM Transactions on Graphics (Presented at ACM SIGGRAPH)}, 2021.

\bibitem[Chen et~al.(2020)Chen, Tagliasacchi, and Zhang]{chen2020bspnet}
Zhiqin Chen, Andrea Tagliasacchi, and Hao Zhang.
\newblock Bsp-net: Generating compact meshes via binary space partitioning.
\newblock \emph{Proceedings of IEEE Conference on Computer Vision and Pattern Recognition (CVPR)}, 2020.

\bibitem[Chen et~al.(2024)Chen, Wang, Wang, Wang, and Liu]{chen2024v3d}
Zilong Chen, Yikai Wang, Feng Wang, Zhengyi Wang, and Huaping Liu.
\newblock V3d: Video diffusion models are effective 3d generators, 2024.

\bibitem[Choi et~al.(2020)Choi, Moon, and Lee]{Choi2020}
Hongsuk Choi, Gyeongsik Moon, and Kyoung~Mu Lee.
\newblock {Pose2Mesh: Graph Convolutional Network for 3D Human Pose and Mesh Recovery from a 2D Human Pose}.
\newblock \emph{Lecture Notes in Computer Science (including subseries Lecture Notes in Artificial Intelligence and Lecture Notes in Bioinformatics)}, 12352 LNCS:\penalty0 769--787, 2020.
\newblock ISSN 16113349.
\newblock \doi{10.1007/978-3-030-58571-6_45}.

\bibitem[Choutas et~al.(2020)Choutas, Pavlakos, Bolkart, Tzionas, and Black]{Choutas2020}
Vasileios Choutas, Georgios Pavlakos, Timo Bolkart, Dimitrios Tzionas, and Michael~J. Black.
\newblock Monocular expressive body regression through body-driven attention.
\newblock In \emph{European Conference on Computer Vision (ECCV)}, 2020.
\newblock URL \url{https://expose.is.tue.mpg.de}.

\bibitem[Corona et~al.(2021)Corona, Pumarola, Aleny{\`a}, Pons-Moll, and Moreno-Noguer]{corona2021smplicit}
Enric Corona, Albert Pumarola, Guillem Aleny{\`a}, Gerard Pons-Moll, and Francesc Moreno-Noguer.
\newblock Smplicit: Topology-aware generative model for clothed people.
\newblock In \emph{CVPR}, 2021.

\bibitem[Deitke et~al.(2023)Deitke, Liu, Wallingford, Ngo, Michel, Kusupati, Fan, Laforte, Voleti, Gadre, VanderBilt, Kembhavi, Vondrick, Gkioxari, Ehsani, Schmidt, and Farhadi]{objaverseXL}
Matt Deitke, Ruoshi Liu, Matthew Wallingford, Huong Ngo, Oscar Michel, Aditya Kusupati, Alan Fan, Christian Laforte, Vikram Voleti, Samir~Yitzhak Gadre, Eli VanderBilt, Aniruddha Kembhavi, Carl Vondrick, Georgia Gkioxari, Kiana Ehsani, Ludwig Schmidt, and Ali Farhadi.
\newblock Objaverse-xl: A universe of 10m+ 3d objects.
\newblock \emph{arXiv preprint arXiv:2307.05663}, 2023.

\bibitem[Du et~al.(2021)Du, Zhang, Yu, Tenenbaum, and Wu]{du2021nerflow}
Yilun Du, Yinan Zhang, Hong-Xing Yu, Joshua~B. Tenenbaum, and Jiajun Wu.
\newblock Neural radiance flow for 4d view synthesis and video processing.
\newblock In \emph{Proceedings of the IEEE/CVF International Conference on Computer Vision}, 2021.

\bibitem[Duggal \& Pathak(2022)Duggal and Pathak]{duggal2022tars3D}
Shivam Duggal and Deepak Pathak.
\newblock Topologically-aware deformation fields for single-view 3d reconstruction.
\newblock \emph{CVPR}, 2022.

\bibitem[Dwivedi et~al.(2021)Dwivedi, Athanasiou, Kocabas, and Black]{Dwivedi2022}
Sai~Kumar Dwivedi, Nikos Athanasiou, Muhammed Kocabas, and Michael~J. Black.
\newblock {Learning to Regress Bodies from Images using Differentiable Semantic Rendering}.
\newblock In \emph{IEEE/CVF International Conference on Computer Vision (ICCV)}, pp.\  11230--11239, 2021.
\newblock ISBN 9781665428125.
\newblock \doi{10.1109/iccv48922.2021.01106}.

\bibitem[Fang et~al.(2022)Fang, Yi, Wang, Xie, Zhang, Liu, Nießner, and Tian]{Fang_2022}
Jiemin Fang, Taoran Yi, Xinggang Wang, Lingxi Xie, Xiaopeng Zhang, Wenyu Liu, Matthias Nießner, and Qi~Tian.
\newblock Fast dynamic radiance fields with time-aware neural voxels.
\newblock In \emph{SIGGRAPH Asia 2022 Conference Papers}, SA ’22. ACM, November 2022.
\newblock \doi{10.1145/3550469.3555383}.
\newblock URL \url{http://dx.doi.org/10.1145/3550469.3555383}.

\bibitem[Feng et~al.(2021)Feng, Choutas, Bolkart, Tzionas, and Black]{Feng2019}
Yao Feng, Vasileios Choutas, Timo Bolkart, Dimitrios Tzionas, and Michael~J. Black.
\newblock Collaborative regression of expressive bodies using moderation.
\newblock In \emph{International Conference on 3D Vision (3DV)}, 2021.

\bibitem[Feng et~al.(2022)Feng, Yang, Pollefeys, Black, and Bolkart]{Feng2022scarf}
Yao Feng, Jinlong Yang, Marc Pollefeys, Michael~J. Black, and Timo Bolkart.
\newblock Capturing and animation of body and clothing from monocular video.
\newblock In \emph{SIGGRAPH Asia 2022 Conference Papers}, SA '22, 2022.

\bibitem[Feng et~al.(2023)Feng, Liu, Bolkart, Yang, Pollefeys, and Black]{Feng2023DELTA}
Yao Feng, Weiyang Liu, Timo Bolkart, Jinlong Yang, Marc Pollefeys, and Michael~J. Black.
\newblock Learning disentangled avatars with hybrid 3d representations.
\newblock \emph{arXiv}, 2023.

\bibitem[Gabeur et~al.(2019)Gabeur, Franco, Martin, Schmid, and Rogez]{gabeur2019moulding}
Valentin Gabeur, Jean-Sebastien Franco, Xavier Martin, Cordelia Schmid, and Gregory Rogez.
\newblock Moulding humans: Non-parametric 3d human shape estimation from single images, 2019.

\bibitem[Gao et~al.(2021)Gao, Saraf, Kopf, and Huang]{Gao-ICCV-DynNeRF}
Chen Gao, Ayush Saraf, Johannes Kopf, and Jia-Bin Huang.
\newblock Dynamic view synthesis from dynamic monocular video.
\newblock In \emph{Proceedings of the IEEE International Conference on Computer Vision}, 2021.

\bibitem[Guan et~al.(2022)Guan, Deng, Wang, and Yang]{guan2022neurofluid}
Shanyan Guan, Huayu Deng, Yunbo Wang, and Xiaokang Yang.
\newblock Neurofluid: Fluid dynamics grounding with particle-driven neural radiance fields, 2022.

\bibitem[Gu{\'e}don \& Lepetit(2023)Gu{\'e}don and Lepetit]{guedon2023sugar}
Antoine Gu{\'e}don and Vincent Lepetit.
\newblock Sugar: Surface-aligned gaussian splatting for efficient 3d mesh reconstruction and high-quality mesh rendering.
\newblock \emph{arXiv preprint arXiv:2311.12775}, 2023.

\bibitem[Guo et~al.(2023)Guo, Yang, Rao, Liang, Wang, Qiao, Agrawala, Lin, and Dai]{guo2023animatediff}
Yuwei Guo, Ceyuan Yang, Anyi Rao, Zhengyang Liang, Yaohui Wang, Yu~Qiao, Maneesh Agrawala, Dahua Lin, and Bo~Dai.
\newblock Animatediff: Animate your personalized text-to-image diffusion models without specific tuning, 2023.

\bibitem[Halimi et~al.(2022)Halimi, Prada, Stuyck, Xiang, Bagautdinov, Wen, Kimmel, Shiratori, Wu, and Sheikh]{halimi2022garment}
Oshri Halimi, Fabian Prada, Tuur Stuyck, Donglai Xiang, Timur Bagautdinov, He~Wen, Ron Kimmel, Takaaki Shiratori, Chenglei Wu, and Yaser Sheikh.
\newblock Garment avatars: Realistic cloth driving using pattern registration, 2022.

\bibitem[He et~al.(2021)He, Xu, Saito, Soatto, and Tung]{He2021}
Tong He, Yuanlu Xu, Shunsuke Saito, Stefano Soatto, and Tony Tung.
\newblock {ARCH++: Animation-Ready Clothed Human Reconstruction Revisited}.
\newblock \emph{Proceedings of the IEEE International Conference on Computer Vision}, pp.\  11026--11036, 2021.
\newblock ISSN 15505499.
\newblock \doi{10.1109/ICCV48922.2021.01086}.

\bibitem[Heming et~al.(2020)Heming, Yu, Hang, Weikai, Dong, Zhangye, Shuguang, and Xiaoguang]{zhu2020deep}
Zhu Heming, Cao Yu, Jin Hang, Chen Weikai, Du~Dong, Wang Zhangye, Cui Shuguang, and Han Xiaoguang.
\newblock Deep fashion3d: A dataset and benchmark for 3d garment reconstruction from single images.
\newblock In \emph{Computer Vision -- ECCV 2020}, pp.\  512--530. Springer International Publishing, 2020.
\newblock ISBN 978-3-030-58452-8.

\bibitem[Ho et~al.(2020)Ho, Jain, and Abbeel]{ho2020denoising}
Jonathan Ho, Ajay Jain, and Pieter Abbeel.
\newblock Denoising diffusion probabilistic models.
\newblock \emph{arXiv preprint arxiv:2006.11239}, 2020.

\bibitem[Hong et~al.(2023)Hong, Zhang, Gu, Bi, Zhou, Liu, Liu, Sunkavalli, Bui, and Tan]{hong2023lrm}
Yicong Hong, Kai Zhang, Jiuxiang Gu, Sai Bi, Yang Zhou, Difan Liu, Feng Liu, Kalyan Sunkavalli, Trung Bui, and Hao Tan.
\newblock Lrm: Large reconstruction model for single image to 3d.
\newblock \emph{arXiv preprint arXiv:2311.04400}, 2023.

\bibitem[Hu et~al.(2023{\natexlab{a}})Hu, Gao, Zhang, Sun, Zhang, and Bo]{hu2023animateanyone}
Li~Hu, Xin Gao, Peng Zhang, Ke~Sun, Bang Zhang, and Liefeng Bo.
\newblock Animate anyone: Consistent and controllable image-to-video synthesis for character animation.
\newblock \emph{arXiv preprint arXiv:2311.17117}, 2023{\natexlab{a}}.

\bibitem[Hu et~al.(2024)Hu, Zhang, Zhang, Zhou, Liu, Zhang, and Nie]{hu2024gaussianavatar}
Liangxiao Hu, Hongwen Zhang, Yuxiang Zhang, Boyao Zhou, Boning Liu, Shengping Zhang, and Liqiang Nie.
\newblock Gaussianavatar: Towards realistic human avatar modeling from a single video via animatable 3d gaussians.
\newblock In \emph{IEEE/CVF Conference on Computer Vision and Pattern Recognition (CVPR)}, 2024.

\bibitem[Hu \& Liu(2023)Hu and Liu]{hu2023gauhuman}
Shoukang Hu and Ziwei Liu.
\newblock Gauhuman: Articulated gaussian splatting from monocular human videos.
\newblock \emph{arXiv preprint arXiv:}, 2023.

\bibitem[Hu et~al.(2023{\natexlab{b}})Hu, Hong, Pan, Mei, Yang, and Liu]{hu2023sherf}
Shoukang Hu, Fangzhou Hong, Liang Pan, Haiyi Mei, Lei Yang, and Ziwei Liu.
\newblock Sherf: Generalizable human nerf from a single image.
\newblock \emph{arXiv preprint arXiv:2303.12791}, 2023{\natexlab{b}}.

\bibitem[Huang et~al.(2023)Huang, Yi, Liu, Wang, Wu, Wang, Lin, Zhang, and Cai]{huang2022elicit}
Yangyi Huang, Hongwei Yi, Weiyang Liu, Haofan Wang, Boxi Wu, Wenxiao Wang, Binbin Lin, Debing Zhang, and Deng Cai.
\newblock One-shot implicit animatable avatars with model-based priors.
\newblock In \emph{IEEE Conference on Computer Vision (ICCV)}, 2023.

\bibitem[Huang et~al.(2020)Huang, Xu, Lassner, Li, and Tung]{Huang2020}
Zeng Huang, Yuanlu Xu, Christoph Lassner, Hao Li, and Tony Tung.
\newblock {ARCH: Animatable Reconstruction of Clothed Humans}.
\newblock \emph{Proceedings of the IEEE Computer Society Conference on Computer Vision and Pattern Recognition}, pp.\  3090--3099, 2020.
\newblock ISSN 10636919.
\newblock \doi{10.1109/CVPR42600.2020.00316}.

\bibitem[Jiang et~al.(2020)Jiang, Zhang, Hong, Luo, Liu, and Bao]{jiang2020bcnet}
Boyi Jiang, Juyong Zhang, Yang Hong, Jinhao Luo, Ligang Liu, and Hujun Bao.
\newblock Bcnet: Learning body and cloth shape from a single image.
\newblock In \emph{European Conference on Computer Vision}. Springer, 2020.

\bibitem[Joo et~al.(2021)Joo, Neverova, and Vedaldi]{Joo2021}
Hanbyul Joo, Natalia Neverova, and Andrea Vedaldi.
\newblock {Exemplar Fine-Tuning for 3D Human Model Fitting Towards In-the-Wild 3D Human Pose Estimation}.
\newblock \emph{Proceedings - 2021 International Conference on 3D Vision, 3DV 2021}, pp.\  42--52, 2021.
\newblock \doi{10.1109/3DV53792.2021.00015}.

\bibitem[Jun \& Nichol(2023)Jun and Nichol]{jun2023shape}
Heewoo Jun and Alex Nichol.
\newblock Shap-e: Generating conditional 3d implicit functions, 2023.

\bibitem[Kanazawa et~al.(2018)Kanazawa, Black, Jacobs, and Malik]{Kanazawa2017}
Angjoo Kanazawa, Michael~J Black, David~W Jacobs, and Jitendra Malik.
\newblock {End-to-End Recovery of Human Shape and Pose}.
\newblock In \emph{Proceedings of the IEEE Computer Society Conference on Computer Vision and Pattern Recognition}, pp.\  7122--7131, 2018.
\newblock ISBN 9781538664209.
\newblock \doi{10.1109/CVPR.2018.00744}.

\bibitem[Kerbl et~al.(2023)Kerbl, Kopanas, Leimk{\"u}hler, and Drettakis]{kerbl3Dgaussians}
Bernhard Kerbl, Georgios Kopanas, Thomas Leimk{\"u}hler, and George Drettakis.
\newblock 3d gaussian splatting for real-time radiance field rendering.
\newblock \emph{ACM Transactions on Graphics}, 42\penalty0 (4), July 2023.
\newblock URL \url{https://repo-sam.inria.fr/fungraph/3d-gaussian-splatting/}.

\bibitem[Kocabas et~al.(2021{\natexlab{a}})Kocabas, Huang, Hilliges, and Black]{Kocabas2021}
Muhammed Kocabas, Chun-Hao~P. Huang, Otmar Hilliges, and Michael~J. Black.
\newblock {PARE}: Part attention regressor for {3D} human body estimation.
\newblock In \emph{Proc. International Conference on Computer Vision (ICCV)}, pp.\  11127--11137, October 2021{\natexlab{a}}.

\bibitem[Kocabas et~al.(2021{\natexlab{b}})Kocabas, Huang, Tesch, M\"uller, Hilliges, and Black]{Kocabas2022}
Muhammed Kocabas, Chun-Hao~P. Huang, Joachim Tesch, Lea M\"uller, Otmar Hilliges, and Michael~J. Black.
\newblock {SPEC}: Seeing people in the wild with an estimated camera.
\newblock In \emph{Proc. International Conference on Computer Vision (ICCV)}, pp.\  11035--11045, October 2021{\natexlab{b}}.

\bibitem[Kocabas et~al.(2023)Kocabas, Chang, Gabriel, Tuzel, and Ranjan]{hugs2024}
Muhammed Kocabas, Rick Chang, James Gabriel, Oncel Tuzel, and Anurag Ranjan.
\newblock Hugs: Human gaussian splats, 2023.
\newblock URL \url{https://arxiv.org/abs/2311.17910}.

\bibitem[Kolotouros et~al.(2019)Kolotouros, Pavlakos, Black, and Daniilidis]{Kolotouros2019}
Nikos Kolotouros, Georgios Pavlakos, Michael Black, and Kostas Daniilidis.
\newblock {Learning to reconstruct 3D human pose and shape via model-fitting in the loop}.
\newblock \emph{Proceedings of the IEEE International Conference on Computer Vision}, 2019-Octob:\penalty0 2252--2261, 2019.
\newblock ISSN 15505499.
\newblock \doi{10.1109/ICCV.2019.00234}.

\bibitem[Kopanas et~al.(2021)Kopanas, Philip, Leimkühler, and Drettakis]{KPLD21}
Georgios Kopanas, Julien Philip, Thomas Leimkühler, and George Drettakis.
\newblock Point-based neural rendering with per-view optimization.
\newblock \emph{Computer Graphics Forum (Proceedings of the Eurographics Symposium on Rendering)}, 40\penalty0 (4), June 2021.
\newblock URL \url{http://www-sop.inria.fr/reves/Basilic/2021/KPLD21}.

\bibitem[Kopanas et~al.(2022)Kopanas, Leimk{\"u}hler, Rainer, Jambon, and Drettakis]{kopanas2022neural}
Georgios Kopanas, Thomas Leimk{\"u}hler, Gilles Rainer, Cl{\'e}ment Jambon, and George Drettakis.
\newblock Neural point catacaustics for novel-view synthesis of reflections.
\newblock \emph{ACM Transactions on Graphics}, 41\penalty0 (6):\penalty0 Article--201, 2022.

\bibitem[Lei et~al.(2023)Lei, Wang, Pavlakos, Liu, and Daniilidis]{lei2023gart}
Jiahui Lei, Yufu Wang, Georgios Pavlakos, Lingjie Liu, and Kostas Daniilidis.
\newblock Gart: Gaussian articulated template models, 2023.
\newblock URL \url{https://arxiv.org/abs/2311.16099}.

\bibitem[Li et~al.(2024{\natexlab{a}})Li, Yao, Xie, and Chen]{li2024gaussianbody}
Mengtian Li, Shengxiang Yao, Zhifeng Xie, and Keyu Chen.
\newblock Gaussianbody: Clothed human reconstruction via 3d gaussian splatting, 2024{\natexlab{a}}.

\bibitem[Li et~al.(2024{\natexlab{b}})Li, Zhou, Zhang, Wei, Hou, and Cheng]{XuanyiLI2024Sora}
Xuanyi Li, Daquan Zhou, Chenxu Zhang, Shaodong Wei, Qibin Hou, and Ming-Ming Cheng.
\newblock Sora generates videos with stunning geometrical consistency.
\newblock \emph{arXiv preprint arXiv: 2402.17403}, 2024{\natexlab{b}}.

\bibitem[Li et~al.(2021)Li, Niklaus, Snavely, and Wang]{li2020neural}
Zhengqi Li, Simon Niklaus, Noah Snavely, and Oliver Wang.
\newblock Neural scene flow fields for space-time view synthesis of dynamic scenes.
\newblock In \emph{Proceedings of the IEEE/CVF Conference on Computer Vision and Pattern Recognition (CVPR)}, 2021.

\bibitem[Li et~al.(2023)Li, Wang, Cole, Tucker, and Snavely]{li2023dynibar}
Zhengqi Li, Qianqian Wang, Forrester Cole, Richard Tucker, and Noah Snavely.
\newblock Dynibar: Neural dynamic image-based rendering.
\newblock In \emph{Proceedings of the IEEE/CVF Conference on Computer Vision and Pattern Recognition (CVPR)}, 2023.

\bibitem[Liu et~al.(2023{\natexlab{a}})Liu, Xu, Jin, Chen, Xu, Su, et~al.]{liu2023one2345}
Minghua Liu, Chao Xu, Haian Jin, Linghao Chen, Zexiang Xu, Hao Su, et~al.
\newblock One-2-3-45: Any single image to 3d mesh in 45 seconds without per-shape optimization.
\newblock \emph{arXiv preprint arXiv:2306.16928}, 2023{\natexlab{a}}.

\bibitem[Liu et~al.(2023{\natexlab{b}})Liu, Wu, Hoorick, Tokmakov, Zakharov, and Vondrick]{liu2023zero1to3}
Ruoshi Liu, Rundi Wu, Basile~Van Hoorick, Pavel Tokmakov, Sergey Zakharov, and Carl Vondrick.
\newblock Zero-1-to-3: Zero-shot one image to 3d object, 2023{\natexlab{b}}.

\bibitem[Liu et~al.(2023{\natexlab{c}})Liu, Huang, Qin, Lin, and Wang]{liu2023animatable}
Yang Liu, Xiang Huang, Minghan Qin, Qinwei Lin, and Haoqian Wang.
\newblock Animatable 3d gaussian: Fast and high-quality reconstruction of multiple human avatars.
\newblock \emph{arXiv preprint arXiv:2311.16482}, 2023{\natexlab{c}}.

\bibitem[Loper et~al.(2015)Loper, Mahmood, Romero, Pons-Moll, and Black]{SMPL2015}
Matthew Loper, Naureen Mahmood, Javier Romero, Gerard Pons-Moll, and Michael~J. Black.
\newblock {SMPL}: A skinned multi-person linear model.
\newblock \emph{ACM Trans. Graphics (Proc. SIGGRAPH Asia)}, 34\penalty0 (6):\penalty0 248:1--248:16, October 2015.

\bibitem[Luiten et~al.(2024)Luiten, Kopanas, Leibe, and Ramanan]{luiten2023dynamic}
Jonathon Luiten, Georgios Kopanas, Bastian Leibe, and Deva Ramanan.
\newblock Dynamic 3d gaussians: Tracking by persistent dynamic view synthesis.
\newblock In \emph{3DV}, 2024.

\bibitem[Melas-Kyriazi et~al.(2023)Melas-Kyriazi, Laina, Rupprecht, and Vedaldi]{melas2023realfusion}
Luke Melas-Kyriazi, Iro Laina, Christian Rupprecht, and Andrea Vedaldi.
\newblock Realfusion: 360deg reconstruction of any object from a single image.
\newblock In \emph{Proceedings of the IEEE/CVF conference on computer vision and pattern recognition}, pp.\  8446--8455, 2023.

\bibitem[Mildenhall et~al.(2020)Mildenhall, Srinivasan, Tancik, Barron, Ramamoorthi, and Ng]{mildenhall2020nerf}
Ben Mildenhall, Pratul~P. Srinivasan, Matthew Tancik, Jonathan~T. Barron, Ravi Ramamoorthi, and Ren Ng.
\newblock Nerf: Representing scenes as neural radiance fields for view synthesis.
\newblock In \emph{ECCV}, 2020.

\bibitem[Moon et~al.(2022)Moon, Choi, and Lee]{Moon2022}
Gyeongsik Moon, Hongsuk Choi, and Kyoung~Mu Lee.
\newblock Accurate 3d hand pose estimation for whole-body 3d human mesh estimation.
\newblock In \emph{Computer Vision and Pattern Recognition Workshop (CVPRW)}, 2022.

\bibitem[Nichol et~al.(2022)Nichol, Jun, Dhariwal, Mishkin, and Chen]{nichol2022pointe}
Alex Nichol, Heewoo Jun, Prafulla Dhariwal, Pamela Mishkin, and Mark Chen.
\newblock Point-e: A system for generating 3d point clouds from complex prompts, 2022.

\bibitem[Park et~al.(2021{\natexlab{a}})Park, Sinha, Barron, Bouaziz, Goldman, Seitz, and Martin-Brualla]{park2021nerfies}
Keunhong Park, Utkarsh Sinha, Jonathan~T. Barron, Sofien Bouaziz, Dan~B Goldman, Steven~M. Seitz, and Ricardo Martin-Brualla.
\newblock Nerfies: Deformable neural radiance fields.
\newblock \emph{ICCV}, 2021{\natexlab{a}}.

\bibitem[Park et~al.(2021{\natexlab{b}})Park, Sinha, Hedman, Barron, Bouaziz, Goldman, Martin-Brualla, and Seitz]{park2021hypernerf}
Keunhong Park, Utkarsh Sinha, Peter Hedman, Jonathan~T. Barron, Sofien Bouaziz, Dan~B Goldman, Ricardo Martin-Brualla, and Steven~M. Seitz.
\newblock Hypernerf: A higher-dimensional representation for topologically varying neural radiance fields.
\newblock \emph{ACM Trans. Graph.}, 40\penalty0 (6), dec 2021{\natexlab{b}}.

\bibitem[Patel et~al.(2020)Patel, Liao, and Pons-Moll]{patel20tailornet}
Chaitanya Patel, Zhouyingcheng Liao, and Gerard Pons-Moll.
\newblock Tailornet: Predicting clothing in 3d as a function of human pose, shape and garment style.
\newblock In \emph{{IEEE} Conference on Computer Vision and Pattern Recognition (CVPR)}. {IEEE}, jun 2020.

\bibitem[Pavlakos et~al.(2019{\natexlab{a}})Pavlakos, Choutas, Ghorbani, Bolkart, Osman, Tzionas, and Black]{SMPLX2019}
Georgios Pavlakos, Vasileios Choutas, Nima Ghorbani, Timo Bolkart, Ahmed~A. Osman, DImitrios Tzionas, and Michael~J. Black.
\newblock {Expressive body capture: 3D hands, face, and body from a single image}.
\newblock In \emph{Proceedings of the IEEE Computer Society Conference on Computer Vision and Pattern Recognition}, volume 2019-June, pp.\  10967--10977, 2019{\natexlab{a}}.
\newblock ISBN 9781728132938.
\newblock \doi{10.1109/CVPR.2019.01123}.

\bibitem[Pavlakos et~al.(2019{\natexlab{b}})Pavlakos, Choutas, Ghorbani, Bolkart, Osman, Tzionas, and Black]{pavlakos2019expressive}
Georgios Pavlakos, Vasileios Choutas, Nima Ghorbani, Timo Bolkart, Ahmed~AA Osman, Dimitrios Tzionas, and Michael~J Black.
\newblock Expressive body capture: 3d hands, face, and body from a single image.
\newblock In \emph{Proceedings of the IEEE/CVF conference on computer vision and pattern recognition}, pp.\  10975--10985, 2019{\natexlab{b}}.

\bibitem[Pons-Moll et~al.(2017)Pons-Moll, Pujades, Hu, and Black]{Pons-Moll:Siggraph2017}
Gerard Pons-Moll, Sergi Pujades, Sonny Hu, and Michael Black.
\newblock Clothcap: Seamless 4d clothing capture and retargeting.
\newblock \emph{ACM Transactions on Graphics, (Proc. SIGGRAPH)}, 36\penalty0 (4), 2017.
\newblock URL \url{http://dx.doi.org/10.1145/3072959.3073711}.
\newblock Two first authors contributed equally.

\bibitem[Poole et~al.(2022)Poole, Jain, Barron, and Mildenhall]{poole2022dreamfusion}
Ben Poole, Ajay Jain, Jonathan~T. Barron, and Ben Mildenhall.
\newblock Dreamfusion: Text-to-3d using 2d diffusion.
\newblock \emph{arXiv}, 2022.

\bibitem[Pumarola et~al.(2020)Pumarola, Corona, Pons-Moll, and Moreno-Noguer]{pumarola2020d}
Albert Pumarola, Enric Corona, Gerard Pons-Moll, and Francesc Moreno-Noguer.
\newblock {D-NeRF: Neural Radiance Fields for Dynamic Scenes}.
\newblock In \emph{Proceedings of the IEEE/CVF Conference on Computer Vision and Pattern Recognition}, 2020.

\bibitem[Qian et~al.(2024{\natexlab{a}})Qian, Mai, Hamdi, Ren, Siarohin, Li, Lee, Skorokhodov, Wonka, Tulyakov, and Ghanem]{Magic123}
Guocheng Qian, Jinjie Mai, Abdullah Hamdi, Jian Ren, Aliaksandr Siarohin, Bing Li, Hsin-Ying Lee, Ivan Skorokhodov, Peter Wonka, Sergey Tulyakov, and Bernard Ghanem.
\newblock Magic123: One image to high-quality 3d object generation using both 2d and 3d diffusion priors.
\newblock In \emph{The Twelfth International Conference on Learning Representations (ICLR)}, 2024{\natexlab{a}}.
\newblock URL \url{https://openreview.net/forum?id=0jHkUDyEO9}.

\bibitem[Qian et~al.(2024{\natexlab{b}})Qian, Wang, Mihajlovic, Geiger, and Tang]{qian20233dgsavatar}
Zhiyin Qian, Shaofei Wang, Marko Mihajlovic, Andreas Geiger, and Siyu Tang.
\newblock 3dgs-avatar: Animatable avatars via deformable 3d gaussian splatting.
\newblock 2024{\natexlab{b}}.

\bibitem[Ren et~al.(2023)Ren, Pan, Tang, Zhang, Cao, Zeng, and Liu]{ren2023dreamgaussian4d}
Jiawei Ren, Liang Pan, Jiaxiang Tang, Chi Zhang, Ang Cao, Gang Zeng, and Ziwei Liu.
\newblock Dreamgaussian4d: Generative 4d gaussian splatting.
\newblock \emph{arXiv preprint arXiv:2312.17142}, 2023.

\bibitem[Rombach et~al.(2021)Rombach, Blattmann, Lorenz, Esser, and Ommer]{rombach2021highresolution}
Robin Rombach, Andreas Blattmann, Dominik Lorenz, Patrick Esser, and Björn Ommer.
\newblock High-resolution image synthesis with latent diffusion models, 2021.

\bibitem[Rong et~al.(2021)Rong, Shiratori, and Joo]{Rong2021}
Yu~Rong, Takaaki Shiratori, and Hanbyul Joo.
\newblock {FrankMocap: A Monocular 3D Whole-Body Pose Estimation System via Regression and Integration}.
\newblock In \emph{Proceedings of the IEEE International Conference on Computer Vision}, volume 2021-Octob, pp.\  1749--1759, 2021.
\newblock ISBN 9781665401913.
\newblock \doi{10.1109/ICCVW54120.2021.00201}.

\bibitem[Saito et~al.(2019)Saito, Huang, Natsume, Morishima, Kanazawa, and Li]{saito2019pifu}
Shunsuke Saito, Zeng Huang, Ryota Natsume, Shigeo Morishima, Angjoo Kanazawa, and Hao Li.
\newblock Pifu: Pixel-aligned implicit function for high-resolution clothed human digitization.
\newblock In \emph{The IEEE International Conference on Computer Vision (ICCV)}, October 2019.

\bibitem[Saito et~al.(2020)Saito, Simon, Saragih, and Joo]{saito2020pifuhd}
Shunsuke Saito, Tomas Simon, Jason Saragih, and Hanbyul Joo.
\newblock Pifuhd: Multi-level pixel-aligned implicit function for high-resolution 3d human digitization.
\newblock In \emph{CVPR}, 2020.

\bibitem[Santesteban et~al.(2019)Santesteban, Otaduy, and Casas]{santesteban2019virtualtryon}
Igor Santesteban, Miguel~A. Otaduy, and Dan Casas.
\newblock {Learning-Based Animation of Clothing for Virtual Try-On}.
\newblock \emph{Computer Graphics Forum (Proc. Eurographics)}, 2019.
\newblock ISSN 1467-8659.
\newblock \doi{10.1111/cgf.13643}.

\bibitem[{Sara Fridovich-Keil and Giacomo Meanti} et~al.(2023){Sara Fridovich-Keil and Giacomo Meanti}, Warburg, Recht, and Kanazawa]{kplanes_2023}
{Sara Fridovich-Keil and Giacomo Meanti}, Frederik~Rahbæk Warburg, Benjamin Recht, and Angjoo Kanazawa.
\newblock K-planes: Explicit radiance fields in space, time, and appearance.
\newblock In \emph{CVPR}, 2023.

\bibitem[Shao et~al.(2023)Shao, Zheng, Tu, Liu, Zhang, and Liu]{shao2023tensor4d}
Ruizhi Shao, Zerong Zheng, Hanzhang Tu, Boning Liu, Hongwen Zhang, and Yebin Liu.
\newblock Tensor4d: Efficient neural 4d decomposition for high-fidelity dynamic reconstruction and rendering.
\newblock In \emph{Proceedings of the IEEE Conference on Computer Vision and Pattern Recognition}, 2023.

\bibitem[Shao et~al.(2024)Shao, Wang, Li, Wang, Lin, Zhang, Fan, and Wang]{SplattingAvatar:CVPR2024}
Zhijing Shao, Zhaolong Wang, Zhuang Li, Duotun Wang, Xiangru Lin, Yu~Zhang, Mingming Fan, and Zeyu Wang.
\newblock {SplattingAvatar: Realistic Real-Time Human Avatars with Mesh-Embedded Gaussian Splatting}.
\newblock In \emph{Computer Vision and Pattern Recognition (CVPR)}, 2024.

\bibitem[Shi et~al.(2023)Shi, Chen, Zhang, Liu, Xu, Wei, Chen, Zeng, and Su]{shi2023zero123plus}
Ruoxi Shi, Hansheng Chen, Zhuoyang Zhang, Minghua Liu, Chao Xu, Xinyue Wei, Linghao Chen, Chong Zeng, and Hao Su.
\newblock Zero123++: a single image to consistent multi-view diffusion base model, 2023.

\bibitem[Szymanowicz et~al.(2023)Szymanowicz, Rupprecht, and Vedaldi]{szymanowicz2023splatter}
Stanislaw Szymanowicz, Christian Rupprecht, and Andrea Vedaldi.
\newblock Splatter image: Ultra-fast single-view 3d reconstruction, 2023.

\bibitem[Tang et~al.(2023{\natexlab{a}})Tang, Ren, Zhou, Liu, and Zeng]{tang2023dreamgaussian}
Jiaxiang Tang, Jiawei Ren, Hang Zhou, Ziwei Liu, and Gang Zeng.
\newblock Dreamgaussian: Generative gaussian splatting for efficient 3d content creation.
\newblock \emph{arXiv preprint arXiv:2309.16653}, 2023{\natexlab{a}}.

\bibitem[Tang et~al.(2024)Tang, Chen, Chen, Wang, Zeng, and Liu]{tang2024lgm}
Jiaxiang Tang, Zhaoxi Chen, Xiaokang Chen, Tengfei Wang, Gang Zeng, and Ziwei Liu.
\newblock Lgm: Large multi-view gaussian model for high-resolution 3d content creation.
\newblock \emph{arXiv preprint arXiv:2402.05054}, 2024.

\bibitem[Tang et~al.(2023{\natexlab{b}})Tang, Wang, Zhang, Zhang, Yi, Ma, and Chen]{Tang_2023_ICCV}
Junshu Tang, Tengfei Wang, Bo~Zhang, Ting Zhang, Ran Yi, Lizhuang Ma, and Dong Chen.
\newblock Make-it-3d: High-fidelity 3d creation from a single image with diffusion prior.
\newblock In \emph{Proceedings of the IEEE/CVF International Conference on Computer Vision (ICCV)}, pp.\  22819--22829, October 2023{\natexlab{b}}.

\bibitem[Tretschk et~al.(2021)Tretschk, Tewari, Golyanik, Zollh\"{o}fer, Lassner, and Theobalt]{tretschk2021nonrigid}
Edgar Tretschk, Ayush Tewari, Vladislav Golyanik, Michael Zollh\"{o}fer, Christoph Lassner, and Christian Theobalt.
\newblock Non-rigid neural radiance fields: Reconstruction and novel view synthesis of a dynamic scene from monocular video.
\newblock In \emph{{IEEE} International Conference on Computer Vision ({ICCV})}. {IEEE}, 2021.

\bibitem[Trevithick \& Yang(2020)Trevithick and Yang]{grf2020}
Alex Trevithick and Bo~Yang.
\newblock Grf: Learning a general radiance field for 3d scene representation and rendering.
\newblock In \emph{arXiv:2010.04595}, 2020.

\bibitem[Turki et~al.(2023)Turki, Zhang, Ferroni, and Ramanan]{turki2023suds}
Haithem Turki, Jason~Y. Zhang, Francesco Ferroni, and Deva Ramanan.
\newblock Suds: Scalable urban dynamic scenes, 2023.

\bibitem[Vidaurre et~al.(2020)Vidaurre, Santesteban, Garces, and Casas]{vidaurre2020virtualtryon}
Raquel Vidaurre, Igor Santesteban, Elena Garces, and Dan Casas.
\newblock {Fully Convolutional Graph Neural Networks for Parametric Virtual Try-On}.
\newblock \emph{Computer Graphics Forum (Proc. SCA)}, 2020.

\bibitem[Wang et~al.(2024)Wang, Ho, Guo, Rong, Grigorev, Song, Zarate, and Hilliges]{wang20244ddress}
Wenbo Wang, Hsuan-I Ho, Chen Guo, Boxiang Rong, Artur Grigorev, Jie Song, Juan~Jose Zarate, and Otmar Hilliges.
\newblock 4d-dress: A 4d dataset of real-world human clothing with semantic annotations.
\newblock In \emph{Proceedings of the IEEE Conference on Computer Vision and Pattern Recognition (CVPR)}, 2024.

\bibitem[Wu et~al.(2023)Wu, Yi, Fang, Xie, Zhang, Wei, Liu, Tian, and Xinggang]{wu20234dgaussians}
Guanjun Wu, Taoran Yi, Jiemin Fang, Lingxi Xie, Xiaopeng Zhang, Wei Wei, Wenyu Liu, Qi~Tian, and Wang Xinggang.
\newblock 4d gaussian splatting for real-time dynamic scene rendering.
\newblock \emph{arXiv preprint arXiv:2310.08528}, 2023.

\bibitem[Xian et~al.(2021)Xian, Huang, Kopf, and Kim]{xian2021space}
Wenqi Xian, Jia-Bin Huang, Johannes Kopf, and Changil Kim.
\newblock Space-time neural irradiance fields for free-viewpoint video.
\newblock In \emph{Proceedings of the IEEE/CVF Conference on Computer Vision and Pattern Recognition (CVPR)}, pp.\  9421--9431, 2021.

\bibitem[Xiang et~al.(2020)Xiang, Prada, Wu, and Hodgins]{Xiang2020}
Donglai Xiang, Fabian Prada, Chenglei Wu, and Jessica Hodgins.
\newblock Monoclothcap: Towards temporally coherent clothing capture from monocular rgb video.
\newblock In \emph{Proceedings of International Conference on 3D Vision (3DV '20)}, pp.\  322 -- 332, November 2020.

\bibitem[Xiang et~al.(2021)Xiang, Prada, Bagautdinov, Xu, Dong, Wen, Hodgins, and Wu]{Xiang_2021}
Donglai Xiang, Fabian Prada, Timur Bagautdinov, Weipeng Xu, Yuan Dong, He~Wen, Jessica Hodgins, and Chenglei Wu.
\newblock Modeling clothing as a separate layer for an animatable human avatar.
\newblock \emph{ACM Transactions on Graphics}, 40\penalty0 (6):\penalty0 1–15, December 2021.
\newblock ISSN 1557-7368.
\newblock \doi{10.1145/3478513.3480545}.
\newblock URL \url{http://dx.doi.org/10.1145/3478513.3480545}.

\bibitem[Xie et~al.(2021)Xie, Wang, Yu, Anandkumar, Alvarez, and Luo]{Xie2021Segformer}
Enze Xie, Wenhai Wang, Zhiding Yu, Anima Anandkumar, Jose~M. Alvarez, and Ping Luo.
\newblock Segformer: Simple and efficient design for semantic segmentation with transformers.
\newblock \emph{CoRR}, abs/2105.15203, 2021.
\newblock URL \url{https://arxiv.org/abs/2105.15203}.

\bibitem[Xie et~al.(2023)Xie, Zong, Qiu, Li, Feng, Yang, and Jiang]{xie2023physgaussian}
Tianyi Xie, Zeshun Zong, Yuxing Qiu, Xuan Li, Yutao Feng, Yin Yang, and Chenfanfu Jiang.
\newblock Physgaussian: Physics-integrated 3d gaussians for generative dynamics.
\newblock \emph{arXiv preprint arXiv:2311.12198}, 2023.

\bibitem[Xiu et~al.(2022)Xiu, Yang, Tzionas, and Black]{xiu2022icon}
Yuliang Xiu, Jinlong Yang, Dimitrios Tzionas, and Michael~J. Black.
\newblock {ICON}: {I}mplicit {C}lothed humans {O}btained from {N}ormals.
\newblock In \emph{Proceedings of the IEEE/CVF Conference on Computer Vision and Pattern Recognition (CVPR)}, pp.\  13296--13306, June 2022.

\bibitem[Xiu et~al.(2023)Xiu, Yang, Cao, Tzionas, and Black]{xiu2023econ}
Yuliang Xiu, Jinlong Yang, Xu~Cao, Dimitrios Tzionas, and Michael~J. Black.
\newblock {ECON: Explicit Clothed humans Optimized via Normal integration}.
\newblock In \emph{Proceedings of the IEEE/CVF Conference on Computer Vision and Pattern Recognition (CVPR)}, June 2023.

\bibitem[Xu et~al.(2019)Xu, Wang, Ceylan, Mech, and Neumann]{Xu2019}
Qiangeng Xu, Weiyue Wang, Duygu Ceylan, Radomir Mech, and Ulrich Neumann.
\newblock Disn: Deep implicit surface network for high-quality single-view 3d reconstruction.
\newblock In H.~Wallach, H.~Larochelle, A.~Beygelzimer, F.~d\textquotesingle Alch\'{e}-Buc, E.~Fox, and R.~Garnett (eds.), \emph{Advances in Neural Information Processing Systems 32}, pp.\  492--502. 2019.

\bibitem[Xu et~al.(2023)Xu, Zhang, Liew, Yan, Liu, Zhang, Feng, and Shou]{xu2023magicanimate}
Zhongcong Xu, Jianfeng Zhang, Jun~Hao Liew, Hanshu Yan, Jia-Wei Liu, Chenxu Zhang, Jiashi Feng, and Mike~Zheng Shou.
\newblock Magicanimate: Temporally consistent human image animation using diffusion model.
\newblock 2023.

\bibitem[Yang et~al.(2023{\natexlab{a}})Yang, Luo, Xiu, Wang, Xu, and Fan]{yang2023dif}
Xueting Yang, Yihao Luo, Yuliang Xiu, Wei Wang, Hao Xu, and Zhaoxin Fan.
\newblock {D-IF: Uncertainty-aware Human Digitization via Implicit Distribution Field}.
\newblock In \emph{Proceedings of the IEEE/CVF international conference on computer vision}, 2023{\natexlab{a}}.

\bibitem[Yang et~al.(2024)Yang, Liu, Zhang, Deng, Huang, and Tan]{yang2024hilo}
Yifan Yang, Dong Liu, Shuhai Zhang, Zeshuai Deng, Zixiong Huang, and Mingkui Tan.
\newblock Hilo: Detailed and robust 3d clothed human reconstruction with high-and low-frequency information of parametric models.
\newblock In \emph{Proceedings of the IEEE/CVF Conference on Computer Vision and Pattern Recognition}, pp.\  10671--10681, 2024.

\bibitem[Yang et~al.(2023{\natexlab{b}})Yang, Cai, Mei, Liu, Chen, Xiao, Wei, Qing, Wei, Dai, Wu, Qian, Lin, Liu, and Yang]{yang2023synbody}
Zhitao Yang, Zhongang Cai, Haiyi Mei, Shuai Liu, Zhaoxi Chen, Weiye Xiao, Yukun Wei, Zhongfei Qing, Chen Wei, Bo~Dai, Wayne Wu, Chen Qian, Dahua Lin, Ziwei Liu, and Lei Yang.
\newblock Synbody: Synthetic dataset with layered human models for 3d human perception and modeling.
\newblock In \emph{Proceedings of the IEEE/CVF International Conference on Computer Vision (ICCV)}, pp.\  20282--20292, October 2023{\natexlab{b}}.

\bibitem[Yang et~al.(2023{\natexlab{c}})Yang, Gao, Zhou, Jiao, Zhang, and Jin]{yang2023deformable3dgs}
Ziyi Yang, Xinyu Gao, Wen Zhou, Shaohui Jiao, Yuqing Zhang, and Xiaogang Jin.
\newblock Deformable 3d gaussians for high-fidelity monocular dynamic scene reconstruction.
\newblock \emph{arXiv preprint arXiv:2309.13101}, 2023{\natexlab{c}}.

\bibitem[Ye et~al.(2023)Ye, Danelljan, Yu, and Ke]{Ye2023gaussian_grouping}
Mingqiao Ye, Martin Danelljan, Fisher Yu, and Lei Ke.
\newblock Gaussian grouping: Segment and edit anything in 3d scenes.
\newblock \emph{arXiv preprint arXiv:2312.00732}, 2023.

\bibitem[Yu et~al.(2023{\natexlab{a}})Yu, Xu, Zhang, Liu, Ye, Wu, Yan, Liang, Chen, Cui, and Han]{yu2023mvimgnet}
Xianggang Yu, Mutian Xu, Yidan Zhang, Haolin Liu, Chongjie Ye, Yushuang Wu, Zizheng Yan, Tianyou Liang, Guanying Chen, Shuguang Cui, and Xiaoguang Han.
\newblock Mvimgnet: A large-scale dataset of multi-view images.
\newblock In \emph{CVPR}, 2023{\natexlab{a}}.

\bibitem[Yu et~al.(2023{\natexlab{b}})Yu, Cheng, Liu, Wu, and Lin]{yu2023monohuman}
Zhengming Yu, Wei Cheng, Xian Liu, Wayne Wu, and Kwan-Yee Lin.
\newblock Monohuman: Animatable human neural field from monocular video.
\newblock In \emph{Proceedings of the IEEE/CVF Conference on Computer Vision and Pattern Recognition}, pp.\  16943--16953, 2023{\natexlab{b}}.

\bibitem[Yuan et~al.(2021)Yuan, Lv, Schmidt, and Lovegrove]{yuan2021star}
Wentao Yuan, Zhaoyang Lv, Tanner Schmidt, and Steven Lovegrove.
\newblock Star: Self-supervised tracking and reconstruction of rigid objects in motion with neural rendering.
\newblock In \emph{Proceedings of the IEEE/CVF Conference on Computer Vision and Pattern Recognition}, pp.\  13144--13152, 2021.

\bibitem[Zakharkin et~al.(2021)Zakharkin, Mazur, Grigorev, and Lempitsky]{Zakharkin_2021_ICCV}
Ilya Zakharkin, Kirill Mazur, Artur Grigorev, and Victor Lempitsky.
\newblock Point-based modeling of human clothing.
\newblock In \emph{Proceedings of the IEEE/CVF International Conference on Computer Vision (ICCV)}, pp.\  14718--14727, October 2021.

\bibitem[Zeng et~al.(2023)Zeng, Wei, Zheng, Zou, Wei, Zhang, and Li]{MakePixelsDance}
Yan Zeng, Guoqiang Wei, Jiani Zheng, Jiaxin Zou, Yang Wei, Yuchen Zhang, and Hang Li.
\newblock Make pixels dance: High-dynamic video generation.
\newblock \emph{arXiv:2311.10982}, 2023.

\bibitem[Zerong et~al.(2021)Zerong, Tao, Yebin, and Qionghai]{zheng2020pamir}
Zheng Zerong, Yu~Tao, Liu Yebin, and Dai Qionghai.
\newblock Pamir: Parametric model-conditioned implicit representation for image-based human reconstruction, 2021.

\bibitem[Zhang et~al.(2023)Zhang, Tian, Zhang, Li, An, Sun, and Liu]{Zhang2022}
Hongwen Zhang, Yating Tian, Yuxiang Zhang, Mengcheng Li, Liang An, Zhenan Sun, and Yebin Liu.
\newblock Pymaf-x: Towards well-aligned full-body model regression from monocular images.
\newblock \emph{IEEE Transactions on Pattern Analysis and Machine Intelligence}, 2023.

\bibitem[Zhu et~al.(2019)Zhu, Zuo, Wang, Cao, and Yang]{zhu2019detailed}
Hao Zhu, Xinxin Zuo, Sen Wang, Xun Cao, and Ruigang Yang.
\newblock Detailed human shape estimation from a single image by hierarchical mesh deformation.
\newblock In \emph{Proceedings of the IEEE/CVF Conference on Computer Vision and Pattern Recognition (CVPR)}, pp.\  4491--4500, 2019.

\bibitem[Zhu et~al.(2024)Zhu, Chen, Dai, Xu, Cao, Yao, Zhu, and Zhu]{zhu2024champ}
Shenhao Zhu, Junming~Leo Chen, Zuozhuo Dai, Yinghui Xu, Xun Cao, Yao Yao, Hao Zhu, and Siyu Zhu.
\newblock Champ: Controllable and consistent human image animation with 3d parametric guidance, 2024.

\end{thebibliography}
\bibliographystyle{neurips_2024}

\clearpage
\section{Appendix}

\subsection{Preliminary}
\label{section:preliminary_extras}

\textbf{3D Gaussian Splatting} utilizes explicit 3D Gaussian points as the core elements for rendering. Each 3D Gaussian point is defined by the function:
\[
G(x) = e^{-\frac{1}{2}(x-\mu)^T \Sigma^{-1} (x-\mu)},
\]
where $\mu$ represents the spatial mean, and $\Sigma$ denotes the covariance matrix. Additionally, each Gaussian is assigned an opacity value $\alpha$ and a view-dependent color $c$, parameterized by spherical harmonic coefficients $f$. During rendering, these 3D Gaussians are projected onto the 2D view plane via a splatting technique. The 2D projection is computed using the projection matrix, while the 2D covariance matrices are approximated as:
$
\Sigma' = J_g W_g \Sigma W_g^T J_g^T,
$
where $W_g$ is the viewing transformation, and $J_g$ is the Jacobian of the affine approximation for perspective projection. The final pixel color is obtained through alpha-blending of $N$ layered 2D Gaussians from front to back 
$
C = \sum_{i \in N} T_i \alpha_i c_i,
$
with
$
T_i = \prod_{j=1}^i (1 - \alpha_j).
$

The opacity $\alpha$ is determined by multiplying $\gamma$ with the contribution of the 2D covariance, derived from $\Sigma'$ and the pixel coordinate in image space. The covariance matrix $\Sigma$ is parameterized using a quaternion $q$ and a 3D scaling vector $v$ to aid in optimization.

\textbf{SMPL-X parameterization}  \cite{SMPLX2019} extends the original SMPL body model \cite{SMPL2015}  by incorporating detailed face and hand deformations to capture more expressive human movements. \textbf{SMPL-X} expands \textbf{SMPL} joint set by including additional joints for facial features, toes and fingers, enabling a more accurate representation of complex body movements.  \textbf{SMPL-X} is defined by a function $M(\beta, \theta, \psi) : \mathbb{R}^{|\beta| \times |\theta| \times |\psi|} \rightarrow \mathbb{R}^{3N}$, where $\theta \in \mathbb{R}^{3K}$ represents the pose (with $K$ being the number of body joints), $\beta \in \mathbb{R}^{|\beta|}$ represents body shape, and $\psi \in \mathbb{R}^{|\psi|}$ captures facial expressions. Further details can be found in \cite{SMPLX2019}.


\subsection{Training details of Identity Encoding loss}
\label{subsection:id_implementation}

To optimize the introduced Identity Encoding of each Gaussian, we render these encoded identity vectors into 2D images in a differentiable manner following \cite{Ye2023gaussian_grouping}.
We adapt the differentiable 3D Gaussian renderer from \cite{kerbl3Dgaussians}, approaching the rendering process similarly to the color optimization using spherical harmonic (SH) coefficients, as described in \cite{kerbl3Dgaussians}. In this method, 3D Gaussian splatting utilizes neural point-based $\alpha'$-rendering \cite{KPLD21,kopanas2022neural}, where the influence weight $\alpha'$ is calculated in 2D for each Gaussian and pixel. Following the approach in \cite{kerbl3Dgaussians}, the influence of all Gaussians on a pixel is computed by sorting them based on depth and blending the $N$ ordered Gaussians that overlap with that pixel:
\begin{equation}
E_{id} = \sum_{i \in \mathcal{N}} e_i \alpha_i \prod_{j=1}^{i-1} (1 - \alpha'_j)
\end{equation}
Here, the rendered 2D mask identity feature $E_{id}$ is the sum of the Identity Encoding $e_{i}$ (of length 15) for each Gaussian, weighted by the Gaussian’s influence factor $\alpha'_i$ on that pixel. The value of $\alpha'_i$ is determined by evaluating a 2D Gaussian with covariance $\Sigma2D$, which is scaled by a learned per-point opacity $\alpha_i$:
\begin{equation}
\Sigma2D = JW\Sigma2D^{3D}W^{T}J^{T} 
\end{equation}
where $\Sigma^{3D}$ is the 3D covariance matrix, $\Sigma2D$ represents the splatted 2D counterpart, $J$ is the Jacobian of the affine approximation for the 3D-to-2D projection, and $W$ is the world-to-camera transformation matrix.

To ensure consistency in the Identity Encoding $e_{i}$ during training, we apply an unsupervised 3D regularization loss. This loss encourages the Identity Encodings of the top $k$-nearest 3D Gaussians to remain close in feature space, promoting spatial consistency. Using the softmax function $F$, we define the KL divergence loss with $m$ sampled points as follows:
\begin{equation}
\mathcal{L}_{3d} = \frac{1}{m} \sum_{j=1}^{m} D_{KL}(P||Q) = \frac{1}{mk} \sum_{j=1}^{m} \sum_{i=1}^{k} F(e_j) \log \left( \frac{F(e_j)}{F'(e_j)} \right)
\end{equation}
Here, $P$ is the sampled Identity Encoding $e$ of a 3D Gaussian, and $Q$ consists of the $k$-nearest neighbors in 3D space, represented as ${e'_1, e'_2, ..., e'_k}$. The total identity encoding loss is then defined as:
\begin{equation}
\mathcal{L}_{id} = \mathcal{L}_{2d} + \mathcal{L}_{3d} 
\end{equation}

\subsection{Implementation details}
\label{subsection:training_details}
The 3D generation experiments were conducted using a single 24GB RTX3090 GPU, while the 4D generation experiments utilized a single 48GB RTX6000 GPU. For the 3D generation process, the SMPL-X fitting was performed with 3000 iterations in 3 minutes, followed by skin color inpainting on SMPL-X Gaussians for 100 iterations in 30 seconds. Reconstruction and disentanglement optimization required 3000 iterations, completed in 12 minutes. In video reconstruction, SMPL-X fitting aligned 14 frames in 6 minutes for in-the-wild videos. The 4D-Dress \cite{wang20244ddress} experiments involved 1000 iterations for clothing deformation over 18 minutes.

\subsection{Visual comparisons to 2D animation methods.}
\begin{figure*}[htbp]
    \centering
    \includegraphics[width=0.95\textwidth ,keepaspectratio]{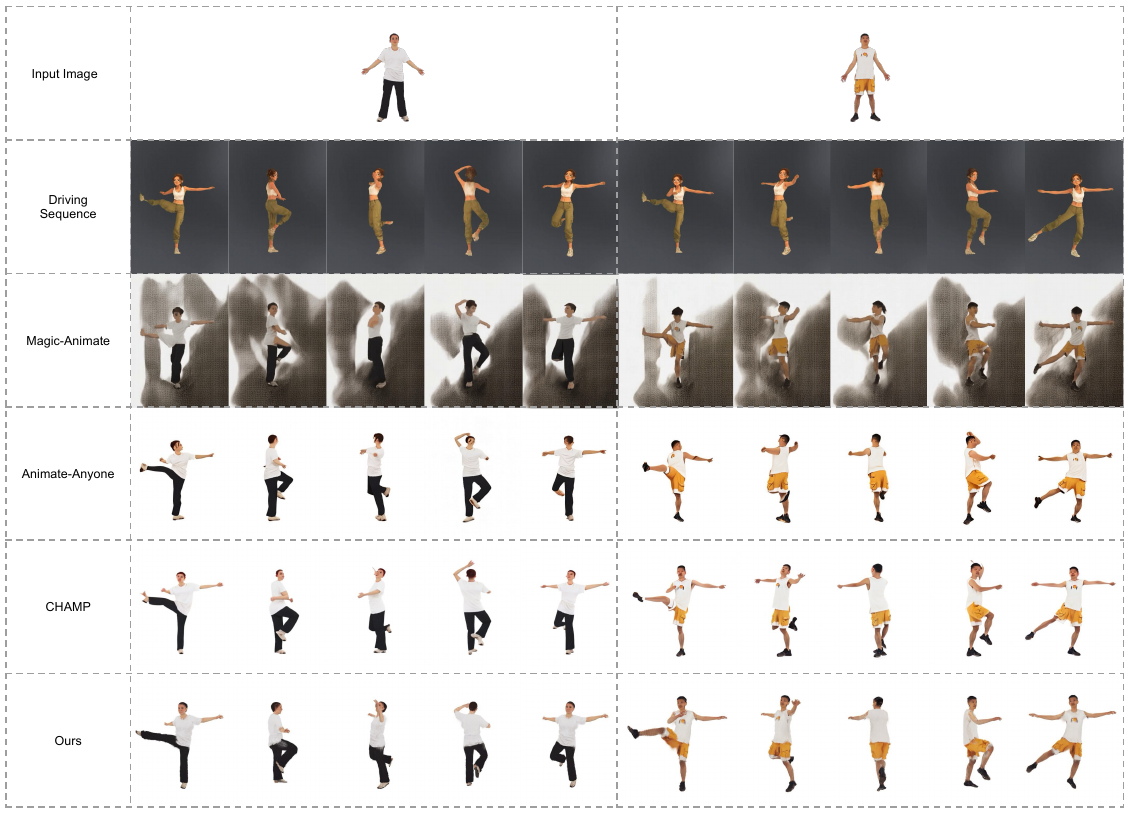}
    \caption{\small \textbf{Comparison to 2D animation methods.} Compared to Magic-Animate and Animate-Anyone, we have better preservation of body shape and details. Compared to CHAMP, we have better geometry and consistency.}
    \label{figure:2d_animation_comparison}
\end{figure*}

\clearpage
\subsection{Failure cases}
\begin{figure*}[htb!]
    \centering
    \includegraphics[width=0.6\linewidth ,keepaspectratio]{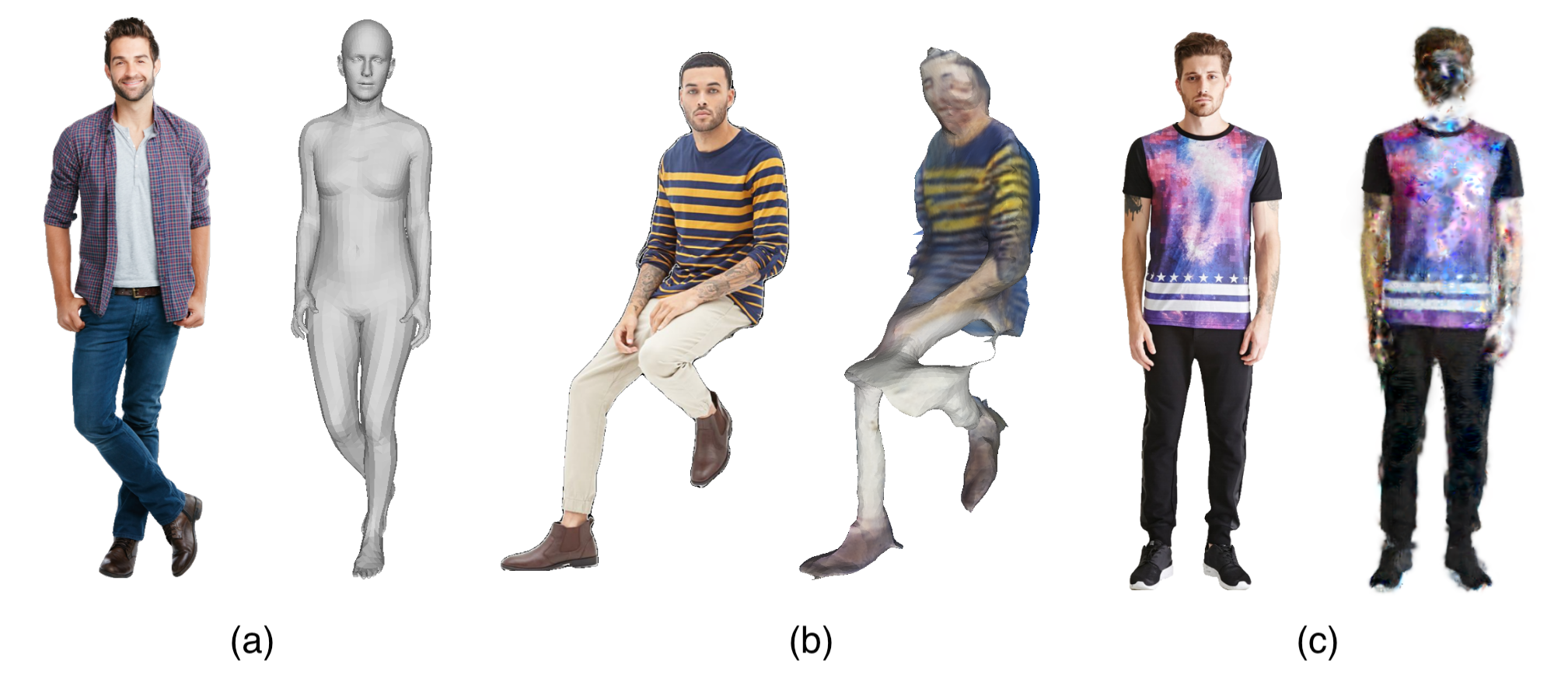}
    \vspace{-5pt}
    \caption{\small \textbf{Failure cases of Disco4D.} (a) Poor SMPL-X estimation  (b) Poor visual hull initialization (c) Misclassification of clothing categories.}
    \label{figure:failure_cases}
\end{figure*}

 Disco4D relies on robust and pixel-aligned SMPL-X estimation, which is still an unsolved problem, especially for challenging poses. In Figure \ref{figure:failure_cases}a, it is difficult to correct the pose with keypoints and segmentation mask due to depth ambiguity. Disco4D occasionally fails for poor visual hull initialization (\ref{figure:failure_cases}b), which is common for difficult poses. Lastly, poor disentanglement is a common problem due to misclassification of clothing category by the segmentation model. This is seen in Figure \ref{figure:failure_cases}c where the arms are wrongly classified under the "top" category.

\subsection{Initialization}
\begin{figure*}[htb!]
    \centering
     \includegraphics[width=0.6\linewidth ,keepaspectratio]{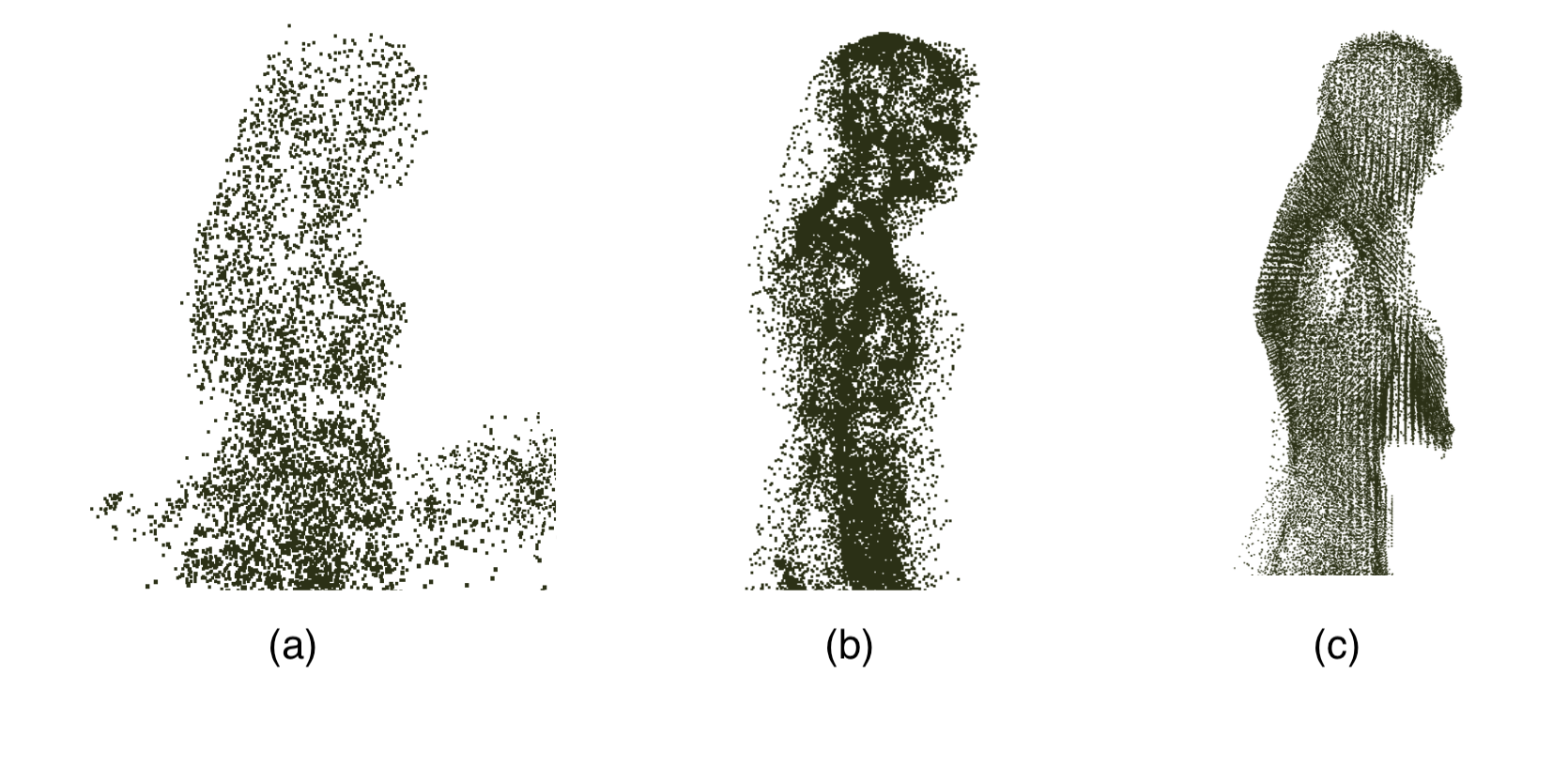}
    \vspace{-10pt}
    \caption{\small \textbf{Ablation of initialization.} (a) Random Initialization (b) SMPL-X Initialization (c) Visual Hull Initialization. }
    \label{figure:ablation_initialisation}
\end{figure*}


\clearpage

\end{document}